\documentclass[runningheads]{llncs}
\usepackage[pdftex]{graphicx}
\usepackage{tikz}
\usepackage{comment}
\usepackage{amsmath,amssymb} 
\usepackage{color}

\usepackage{subfiles}
\usepackage{multirow}
\usepackage{amsmath}
\usepackage{booktabs}  
\usepackage{subfigure}
\usepackage{wrapfig}
\usepackage{xcolor}
\definecolor{mygreen}{RGB}{0, 153, 0}

\usepackage[accsupp]{axessibility}  
\newcommand{\Paragraph}[1]{\vspace{0.1mm} \noindent \textbf{#1} \hspace{0mm}}
\usepackage{amsmath}
\DeclareMathOperator*{\argmax}{arg\,max}

\usepackage{cite}
\usepackage[pagebackref,breaklinks,colorlinks]{hyperref}

\begin{document}
\pagestyle{headings}
\mainmatter
\def\ECCVSubNumber{2887}  

\title{ Controllable and Guided Face Synthesis for Unconstrained Face Recognition}

\titlerunning{CFSM for Unconstrained Face Recognition}
%
\author{Feng Liu, Minchul Kim, Anil Jain, and Xiaoming Liu}
\authorrunning{F. Liu et al.}
\institute{Michigan State University, Computer Science \& Engineering\\
\email{\{liufeng6,kimminc2,jain,liuxm\}@msu.edu}}

\maketitle

\begin{abstract}

Although significant advances have been made in face recognition (FR), FR in unconstrained environments remains challenging due to the domain gap between the semi-constrained training datasets and unconstrained testing scenarios.
To address this problem, we propose a controllable face synthesis model (CFSM) that can mimic the distribution of target datasets in a style latent space. 
CFSM learns a linear subspace with orthogonal bases in the style latent space with precise control over the diversity and degree of synthesis. 
Furthermore, the pre-trained synthesis model can be guided by the FR model, making the resulting images more beneficial for FR model training.
Besides, target dataset distributions are characterized by the learned orthogonal bases, which can be utilized to measure the distributional similarity among face datasets.
Our approach yields significant performance gains on unconstrained benchmarks, such as IJB-B, IJB-C, TinyFace and IJB-S (${+}5.76\%$ Rank$1$). Code is available at \url{http://cvlab.cse.msu.edu/project-cfsm.html}.

\end{abstract}


\section{Introduction}

Face recognition (FR) is now one of the most well-studied problems in the area of computer vision and pattern recognition. 
The rapid progress in face recognition accuracy can be attributed to developments in deep neural network models~\cite{he2016deep,huang2017densely,simonyan2014very,tan2019efficientnet}, sophisticated design of loss functions~\cite{deng2019arcface,liu2017sphereface,wang2018cosface,wen2016discriminative,wang2017normface,wang2018additive,zheng2018ring,meng2021magface,wang2020mis,liu2019adaptiveface,huang2020curricularface,sun2020circle}, and large-scale training datasets, \emph{e.g.}, MS-Celeb-1M~\cite{guo2016ms} and WebFace260M~\cite{zhu2021webface260m}.

Despite this progress, state-of-the-art (SoTA) FR models do not work well on real-world surveillance imagery (unconstrained) due to the domain shift issue, that is, the large-scale training datasets (semi-constrained) obtained via web-crawled celebrity faces lack in-the-wild variations, such as inherent sensor noise, low resolution, motion blur, turbulence effect, \emph{etc}. 
For instance, $1{:}1$ verification accuracy reported by one of the SoTA models~\cite{shi2019probabilistic} on unconstrained IJB-S~\cite{kalka2018ijb} dataset is about $30\%$ lower than on semi-constrained LFW~\cite{huang2008labeled}.
A potential remedy to such a performance gap is to assemble a large-scale unconstrained face dataset. 
However, constructing such a training dataset with tens of thousands of subjects is prohibitively difficult with high manual labeling cost.


An alternative solution is to develop facial image generation models that can synthesize face images with desired properties.
Face translation or synthesis using GANs~\cite{karras2019style,karras2020analyzing,choi2018stargan,karras2017progressive,yang2021gan,wang2021towards} or $3$D face reconstruction~\cite{thies2016face2face,rossler2019faceforensics++,tran2018nonlinear,kim2018deep,zhu2016face,riggable-3d-face-reconstruction-via-in-network-optimization,face-relighting-with-geometrically-consistent-shadows} has been well studied in photo-realistic image generation. 
However, most of these methods mainly focus on face image restoration or editing, and hence do not lead to better face recognition accuracies. 
A recent line of research~\cite{kortylewski2019analyzing,shi2020towards,qiu2021synface,trigueros2021generating} adopts disentangled face synthesis~\cite{tewari2020stylerig,deng2020disentangled,tran2017disentangled}, which can provide control over explicit facial properties (pose, expression and illumination) for generating additional synthetic data for varied training data distributions.
However, the hand-crafted categorization of facial properties and lack of design for cross-domain translation limits their generalizability to challenging testing scenarios. 
Shi \emph{et al.}~\cite{shi2021boosting} propose to use an unlabeled dataset to boost unconstrained face recognition. However, all of the previous methods can be considered as performing \textit{blind} data augmentation, {\it i.e.},  without the feedback of the FR model, which is required to provide critical information for improving the FR performance.

\begin{figure}[t]
\centering
\includegraphics[width=1\textwidth]{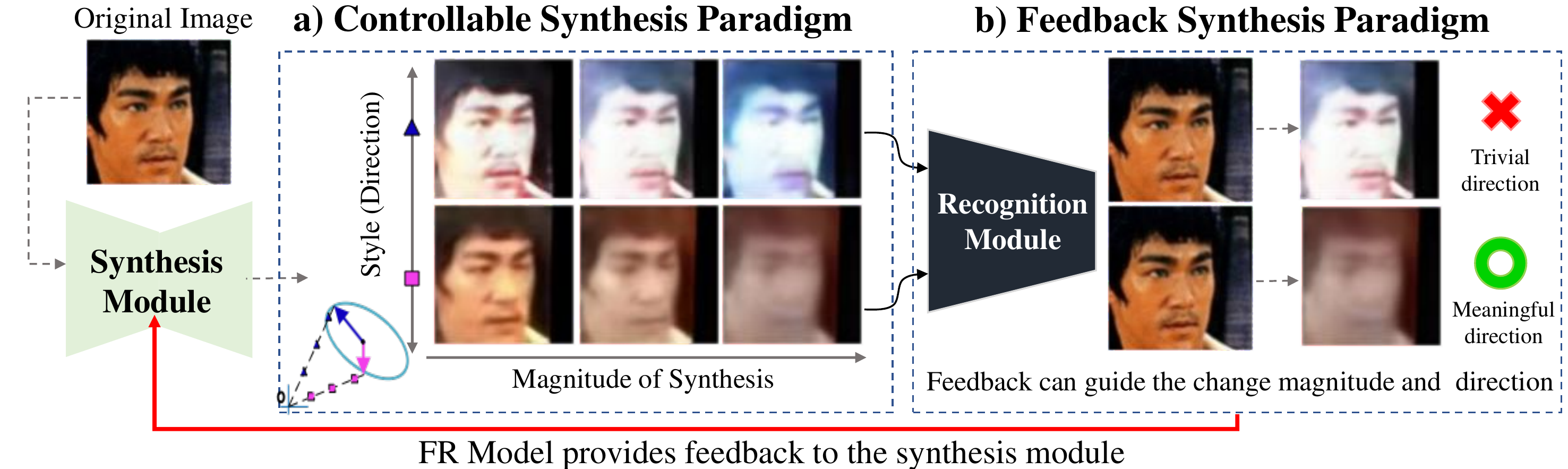}
\vspace{-2mm}
\caption{ (a) Given an input face image, our controllable face synthesis model (CFSM) enables precise control of the direction and magnitude of the targeted styles in the generated images. The latent style has both the direction and the magnitude, where the direction linearly combines the learned bases to control the \textit{type} of style, while the magnitude controls the \textit{degree} of style. 
(b) CFSM can incorporate the feedback provided by the FR model to generate synthetic training data that can benefit the FR model training and improve generalization to the unconstrained testing scenarios.}
\vspace{-2mm}
\label{fig:teaser}
\end{figure}

In Fig.~\ref{fig:teaser}, we show the difference between a blind and feedback-based face synthesis paradigm. For blind face synthesis, the FR model does not take part in the synthesis process, so there is no guidance from the FR model to avoid trivial synthesis. With feedback from the FR model, as in Fig.~\ref{fig:teaser} (b), synthesized images can be more relevant to increasing the FR performance. Therefore, it is the goal of our paper to allow the FR model to guide the face synthesis towards creating synthetic datasets that can improve the FR performance.


It is not trivial to incorporate the signal from the FR model, as the direct manipulation of an input image towards decreasing the FR loss results in adversarial images that are not analogous to the real image distribution~\cite{goodfellow2014explaining}. We thus propose to learn manipulation in the subspace of the \textit{style space} of the target properties, so that the control can be accomplished 1) in low-dimensions, 2) semantically meaningful along with various quality factors. 

In light of this, this paper aims to answer these three questions: 

\emph{1. Can we learn a face synthesis model that can discover the styles in the target unconstrained data, which enables us to precisely control and increase the diversity of the labeled training samples?}

\emph{2. Can we incorporate the feedback provided by the FR model in generating synthetic training data, towards facilitating FR model training?}

\emph{3. Additionally, as a by-product of our proposed style based synthesis, can we model the distribution of a target dataset, so that it allows us to quantify the distributional similarity among face datasets?}

\noindent Towards this end, we propose a face synthesis model that is 1) controllable in the synthesis process and 2) guided in the sense that the sample generation is aided by the signal from the FR model.
Specifically, given a labeled training sample set, our controllable face synthesis model (CFSM) is trained to discover different attributes of the unconstrained target dataset in a style latent space.
To learn the explicit degree and direction that control the styles in an unsupervised manner, we embed one linear subspace model with orthogonal bases into the style latent space. 
Within generative adversarial training, the face synthesis model seeks to capture the principal variations of the data distribution, and the style feature magnitude controls the degree of manipulation in the synthesis process.   

More importantly, to extract the feedback of the FR model, we apply adversarial perturbations (FGSM) in the learned style latent space to guide the sample generation. 
This feedback is rendered meaningful and efficient because the manipulation is in the low dimensional style space as opposed to in the high dimensional image space.
With the feedback from the FR model, the synthesized images are more beneficial to the FR performance, leading to significantly improved generalization capabilities of the FR models trained with them. 
It is worth noting that our pre-trained synthesis models could be a plug-in to any SoTA FR model. 
Unlike the conventional face synthesis models that focus on high quality realistic facial images, our face synthesis module is a conditional mapping from one image to a set of style shifted images that match the distribution of the target unconstrained dataset towards boosting its FR performance.

Additionally, the learned orthogonal bases characterize the target dataset distribution that could be utilized to quantify distribution similarity between datasets. 
The quantification of datasets has broad impact on various aspects. 
For example, knowing the dataset distribution similarity could be utilized to gauge the expected performance of FR systems in new datasets. 
Likewise, given a choice of various datasets to train an FR model, one can find one closest to the testing scenario of interest. 
Finally, when a new face dataset is captured in the future, we may also access its similarity to existing datasets in terms of styles, in addition to typical metrics such as number of subjects, demographics, etc.

In summary, the contributions of this work include:

$\diamond$ We show that a controllable face synthesis model with linear subspace style representation can generate facial images of the target dataset style, with precise control in the magnitude and type of style.

$\diamond$ We show that FR model performance can be greatly increased by synthesized images when the feedback of the FR model is used to optimize the latent style coefficient during image synthesis.

$\diamond$ Our learned linear subspace model can characterize the target dataset distribution for quantifying the distribution similarity between face datasets. 

$\diamond$ Our approach yields significant performance gains on unconstrained face recognition benchmarks, such as IJB-B, IJB-C, IJB-S and TinyFace.

\section{Prior Work}\label{sec:prior}

\Paragraph{Controllable Face Synthesis} With the remarkable ability of GANs~\cite{goodfellow2014generative}, face synthesis has seen rapid developments, such as 
StyleGAN~\cite{karras2019style} and its variations~\cite{karras2020analyzing,Karras2020ada,karras2021alias} which can generate high-fidelity face images from random noises.
Lately, GANs have seen widespread use in face image translation or manipulation~\cite{hu2018disentangling,deng2020disentangled,xiao2018elegant,pumarola2018ganimation,sun2019single,choi2018stargan,lin2018conditional,bulat2018learn}. These methods typically adopt an encoder-decoder/generator-discriminator paradigm where the encoder embeds images into disentangled latent representations characterizing different face properties. 
Another line of works incorporates $3$D prior (\emph{i.e.}, $3$DMM~\cite{blanz1999morphable}) into GAN for $3$D-controllable face synthesis~\cite{shen2018facefeat,kim2018deep,deng2018uv,gecer2018semi,geng20193d,piao2019semi,nguyen2019hologan,most-gan-3d-morphable-stylegan-for-disentangled-face-image-manipulation}. Also, EigenGAN ~\cite{he2021eigengan} introduces the linear subspace model into each generator layer, which enables to discover layer-wise interpretable variations. 
Unfortunately, these methods mainly focus on high-quality face generation or editing on pose, illumination and age, which has a well-defined semantic meaning. However, style or domain differences are hard to be factorized at the semantic level. Therefore, we utilize learned bases to cover unconstrained dataset attributes, such as resolution, noise, \emph{etc}. 
%

\Paragraph{Face Synthesis for Recognition} 
Early attempts exploit disentangled face synthesis to generate additional synthetic training data for either reducing the negative effects of dataset bias in FR ~\cite{kortylewski2019analyzing,shi2020towards,qiu2021synface,trigueros2021generating,ruiz2020morphgan} or more efficient training of pose-invariant FR models~\cite{tran2017disentangled,yin2017towards,zhao20183d}, resulting in increased FR accuracy. 
However, these models only control limited face properties, such as pose, illumination and expression, which are not adequate for bridging the domain gap between the semi-constrained and unconstrained face data.
The most pertinent study to our work is~\cite{shi2021boosting}, which proposes to generalize face representations with auxiliary unlabeled data. Our framework differs in two aspects: 
i) our synthesis model is precisely-controllable in the style latent space, in both magnitude and direction, and 
ii) our synthesis model incorporates \emph{guidance} from the FR model, which significantly improves the generalizability to unconstrained FR.

\Paragraph{Domain Generalization and Adaptation}
Domain Generalization (DG) aims to make DNN perform well on unseen domains~\cite{muandet2013domain,ghifary2015domain,ghifary2016scatter,motiian2017unified,li2018domain}. 
Conventionally, for DG, few labeled samples are provided for the target domain to generalize. 
Popular DG methods  utilize auxiliary losses such as Maximum
Mean Discrepancy or domain adversarial loss to learn a shared feature space across multiple source domains~\cite{muandet2013domain,li2018domain,ghifary2016scatter}. 
In contrast, our method 
falls into the category of Unsupervised Domain Adaptation where adaptation is achieved by adversarial loss, contrastive loss or image translation, and learning a shared feature space that works for both the original and target domains~\cite{ganin2015unsupervised,tzeng2017adversarial,saito2018maximum,kang2019contrastive,murez2018image,nam2021reducing}. 
Our method augments data resembling the target domain with unlabeled images. 

\Paragraph{Dataset Distances}
It is important to characterize and contrast datasets in computer vision research. 
In recent years, various notions of dataset similarity have been proposed~\cite{david,Mansour,alvarez2020geometric}. Alpha-distance and discrepancy distance~\cite{david,Mansour} measures a dissimilarity that depends on a loss function and the predictor. To avoid the dependency on the label space,~\cite{alvarez2020geometric} proposes OT distance, an optimal transport distance in the feature space. However, it still depends on the ability of the predictor to create a separable feature space across domains. Moreover, a feature extractor trained on one domain may not predict the resulting features as separable in a new domain. In contrast, we propose to utilize the learned linear bases for latent style codes, which are optimized for synthesizing images in target domains, to measure the dataset distance. The style-based distance has the benefit of not being dependent on the feature space or the label space. 

\section{Proposed Method}

\begin{figure}[t]
\centering
\includegraphics[width=12.0cm]{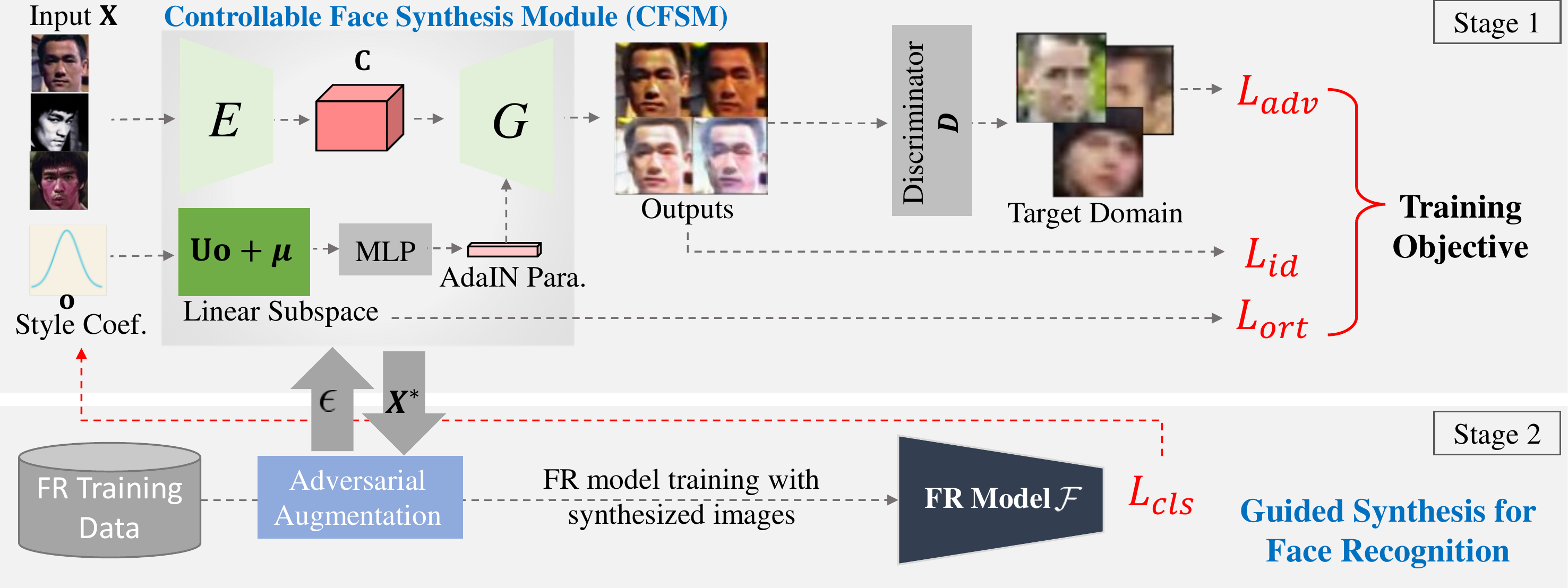}
\caption{Overview of the proposed method. Top (Stage 1): Pipeline for training the controllable face synthesis module that mimics the distribution of the target domain. $\mathcal{L}_{adv}$ ensures target domain similarity,  $\mathcal{L}_{id}$  enforces the magnitude of $\mathbf{o}$ to control the degree of synthesis, and  $\mathcal{L}_{ort}$ factorizes the target domain style with linear bases.  
Bottom (Stage 2): Pipeline for using the pre-trained face synthesis module for the purpose of training an FR model. The synthesis module works as an augmentation to the training data. We adversarially update $\mathbf{o}$ to maximize $\mathcal{L}_{cls}$ of a given FR model.}
\label{fig:synthesis}
\end{figure}

\subsection{Controllable Face Synthesis Model}
Generally, for face recognition model training, we are given a labeled semi-constrained dataset that consists of $n$ face images $\mathcal{X}={\mathbf{\{X}\}_{i=1}^{n}}$ and the corresponding identity labels. 
Meanwhile, similar to the work in~\cite{shi2021boosting}, we assume the availability of an unlabeled target face dataset with $m$ images $\mathcal{Y}={\mathbf{\{Y}\}_{i=1}^{m}}$, which contains a large variety of unconstrained factors. 
Our goal is to learn a style latent space where the face synthesis model, given an input image from the semi-constrained dataset, can generate new face images of the same subject, whose style is similar to the target dataset. 
Due to the lack of corresponding images, we seek an unsupervised algorithm that can learn to translate between domains without paired input-output examples.  
In addition, we hope this face synthesis model has explicit dimensions to control the unconstrained attributes.

\emph{Our face synthesis model is not designed to translate the intrinsic properties between faces, i.e., pose, identity or expression.} It is designed to focus on capturing the unconstrained imaging environment factors in unconstrained face images, such as noise, low resolution, motion blur, turbulence effects, etc. These variations are not present in large-scale labeled training data for face recognition.

\Paragraph{Multimodal Image Translation Network.}
We adopt a multimodal image-to-image translation network~\cite{huang2018multimodal,lee2018diverse} to discover the underlying style distribution in the target domain. Specifically, as shown in Fig.~\ref{fig:synthesis}, our face synthesis generator consists of an encoder $E$ and a decoder $G$. Given an input image $\mathbf{X}\in\mathbb{R}^{W\times H\times3}$, the encoder first extracts its content features $\mathbf{C}=E(\mathbf{X})$.
Then, the decoder generates the output image $\hat{\mathbf{X}}\in\mathbb{R}^{W\times H\times3}$, conditioned on both the content features and a random style latent code $\mathbf{z}\in\mathcal{Z}^{d}$: $\hat{\mathbf{X}}=G(\mathbf{C}, \mathbf{z})$. Here, the style code $\mathbf{z}$ is utilized to control the style of the output image.

Inspired by recent works that use affine transformation parameters in normalization layers to represent image styles~\cite{huang2017arbitrary,dumoulin2016learned,karras2019style,huang2018multimodal}, we equip the residual blocks in the decoder $D$ with Adaptive Instance Normalization (AdaIN) layers~\cite{huang2017arbitrary}, whose parameters are dynamically generated by a multilayer perceptron (MLP) from the style code $\mathbf{z}$. Formally, the decoder process can be presented as
\begin{equation}
    \hat{\mathbf{X}}=G(\mathbf{C}, \textup{MLP}(\mathbf{z})).
    \label{eqn:dec}
\end{equation}
It is worth noting that such $G$ can model continuous distributions which enables us to generate multimodal outputs from a given input.

We employ one adversarial discriminator $D$ to match the distribution of the synthesized images to the target data distribution; images generated by the model should be indistinguishable from real images in the target domain. The discriminator loss can be described as:
\begin{equation}
    \mathcal{L}_{D} = -\mathbb{E}_{ \mathbf{Y}\sim \mathcal{Y}}[\textup{log}(D(\mathbf{Y}))] - \mathbb{E}_{\mathbf{X}\sim \mathcal{X},\mathbf{z}\sim \mathcal{Z} }[\textup{log}(1-D(\hat{\mathbf{X}}))].
    \label{eqn:dis_loss}
\end{equation}
The adversarial loss for the generator (including $E$ and $G$) is then defined as:
\begin{equation}
    \mathcal{L}_{adv} = -\mathbb{E}_{\mathbf{X}\sim \mathcal{X},\mathbf{z}\sim \mathcal{Z}}[\textup{log}(D(\hat{\mathbf{X}}))].
    \label{eqn:gen_loss}
\end{equation}

\Paragraph{Domain-Aware Linear Subspace Model.}
To enable precise control of the targeted face properties, achieving flexible image generation, we propose to embed a linear subspace model with orthogonal bases into the style latent space. As illustrated in Fig.~\ref{fig:synthesis},
a random style coefficient $\mathbf{o}\sim \mathcal{N}_{q}(\mathbf{0},\mathbf{I})$ can be used to linearly combine the bases and form a new style code $\mathbf{z}$, as in
\begin{equation}
    \mathbf{z} = \mathbf{U}\mathbf{o}+ \boldsymbol\mu,
    \label{eqn:pca}
\end{equation}
where $\mathbf{U}=[\mathbf{u}_{1}, \cdots, \mathbf{u}_{q}]\in\mathbb{R}^{d\times q}$ is the orthonormal basis of the subspace. $\boldsymbol\mu\in \mathbb{R}^{d}$ denotes the mean style. This equation relates a $q$-dimensional coefficient $\mathbf{o}$ to a corresponding $d$-dimensional style vector ($q<<d$) by an affine transformation and translation. During training, both $\mathbf{U}$ and $\boldsymbol\mu$ are learnable parameters. The entire bases $\mathbf{U}$ are optimized with the orthogonality constraint~\cite{he2021eigengan}: $\mathcal{L}_{ort}=|\mathbf{U}^{\textup{T}}\mathbf{U}-\mathbf{I}|_1$, where $\mathbf{I}$ is an identity matrix.

The isotropic prior distribution of $\mathbf{o}$ does not indicate which directions are useful. However, with the help of the subspace model, each basis vector in $\mathbf{U}$ identifies a latent direction that allows control over target image attributes that vary from straightforward high-level face properties. This mechanism is algorithmically simple, yet leads to effective control without requiring ad-hoc supervision. Accordingly, Eqn.~\ref{eqn:dec} can be updated as $\hat{\mathbf{X}}=G(\mathbf{C}, \textup{MLP}(\mathbf{U}\mathbf{o}+ \boldsymbol\mu))$.  

\noindent \Paragraph{Magnitude of the Style Coefficient and Identity Preservation.} Although the adversarial learning (Eqn.~\ref{eqn:dis_loss} and~\ref{eqn:gen_loss}) could encourage the face synthesis module to characterize the attributes in the target data, it cannot ensure the identity information is maintained in the output face image. Hence, the cosine similarity $S_{C}$ between the face feature vectors $f(\mathbf{X})$ and $f(\hat{\mathbf{X}})$ is used to enforce identity preservation: $\mathcal{L}_{id}{=}1{-} S_{C}(f(\mathbf{X}),f(\hat{\mathbf{X}}))$, where $f(\cdot)$ represents a pre-trained feature extractor, \emph{i.e.,} ArcFace~\cite{deng2019arcface} in our implementation.

Besides identifying the meaningful latent direction, we continue to explore the property of the magnitude $a=||\mathbf{o}||$ of the style coefficient. We expect the magnitude can measure the degree of identity-preservation in the synthesized image $\hat{\mathbf{X}}$. In other words, $S_{C}(f(\mathbf{X}),f(\hat{\mathbf{X}}))$ monotonically increases when the magnitude $a$ is decreased. To realize this goal, we re-formulate the identity loss:
\begin{equation}
    \mathcal{L}_{id} = \left \| \left( 1- S_{C}(f(\mathbf{X}),f(\hat{\mathbf{X}})) \right)-g(a) \right \|^2_{2},
    \label{equ:id_loss}
\end{equation}
where $g(a)$ is a function with respect to $a$. We assume the magnitude $a$ is bounded in $[l_a, u_a]$. In our implementation, we define $g(a)$ as a linear function on $[l_a, u_a]$ with $g(l_a) = l_m$, $g(u_a) = u_m$: $g(a)=(a-l_a)\frac{u_m-l_m}{u_a-l_a}+l_m$.

By simultaneously learning the direction and magnitude of the style latent coefficients, our model becomes \emph{precisely controllable} in capturing the variability of faces in the target domain. 
To our knowledge, this is the first method which is able to explore the complete set of two properties associated with the style, namely direction and magnitude, in unsupervised multimodal face translation.

\Paragraph{Model learning.}
The total loss for the generator (including encoder $E$, decoder $G$ and domain-aware linear subspace model), with weights $\lambda_{i}$, is 
\begin{equation}
    \mathcal{L}_{\mathcal{G}} = \lambda_{adv} \mathcal{L}_{adv} + \lambda_{ort}  \mathcal{L}_{ort} + \lambda_{id} \mathcal{L}_{id}.
    \label{equ:overall_loss}
\end{equation}

\subsection{Guided Face Synthesis for Face Recognition}
\label{sec:guide}

In this section, we introduce how to incorporate the \emph{pre-trained} face synthesis module into deep face representation learning, enhancing the generalizability to unconstrained FR. It is effectively addressing, \emph{which synthetic images, when added as an augmentation to the data, will increase the performance of the learned FR model in the unconstrained scenarios?}

Formally, the FR model is trained to learn a mapping $\mathcal{F}$, such that $\mathcal{F}(\mathbf{X})$ is discriminative for different subjects. If $\mathcal{F}$ is only trained on the domain defined by semi-constrained $ \mathcal{X}$, it does not generalize well to unconstrained scenarios. 
However, $\mathbf{X}$ with identity label $l$ in a training batch may be augmented with a random style coefficient $\mathbf{o}$ to produce a synthesized image $\hat{\mathbf{X}}$ with CFSM.

However, such data synthesis with random style coefficients may generate either extremely easy or hard samples, which may be redundant or detrimental to the FR training.
To address this issue, we introduce an adversarial regularization strategy to \emph{guide} the data augmentation process, so that the face synthesis module is able to generate meaningful samples for the FR model.
Specifically, for a given pre-trained CFSM, we apply adversarial perturbations in the learned style latent space, in the direction of maximizing the FR model loss. 
Mathematically, given the perturbation budget $\epsilon$, the adversary tries to find a style latent perturbation $\boldsymbol\delta\in\mathbb{R}^d$ to maximize the classification loss function $\mathcal{L}_{cla}$:
\begin{equation}
\begin{aligned}
    \boldsymbol\delta^{*}=\argmax_{||\boldsymbol\delta||_{\infty }<\epsilon}\mathcal{L}_{cla}\left(\mathcal{F}(\mathbf{X}^{*}),l\right), \text{where} \;
    \mathbf{X}^{*}\! =\! G(E(\mathbf{X}),\textup{MLP}(\mathbf{U}(\mathbf{o}+\boldsymbol\delta)+\boldsymbol\mu)).
\end{aligned}
    \label{eqn:inter_max}
\end{equation}

Here, $\mathbf{X}^{*}$ denotes the perturbed synthesized image. $\mathcal{L}_{cla}$ could be any \\
classification-based loss, \emph{e.g.,} popular angular margin-based loss, ArcFace~\cite{deng2019arcface} in our implementation. 
In this work, for efficiency, we adopt the one-step Fast Gradient Sign Method (FGSM)~\cite{goodfellow2014explaining} to obtain $\boldsymbol\delta^{*}$ and subsequently update $\mathbf{o}$:

\begin{equation}
\begin{aligned}
    \mathbf{o}^{*} = \mathbf{o} + \boldsymbol\delta^{*}, \quad \boldsymbol\delta^{*}=\epsilon\cdot \text{sgn}\left( \nabla_{\mathbf{z}}\mathcal{L}_{cla}\left(\mathcal{F}(\mathbf{X}^{*}),l\right)\right), 
\end{aligned}
    \label{eqn:perb}
\end{equation}
\begin{wrapfigure}[15]{r}{0.5\textwidth}
\vspace{-8mm}
\centering
\includegraphics[width=0.5\textwidth]{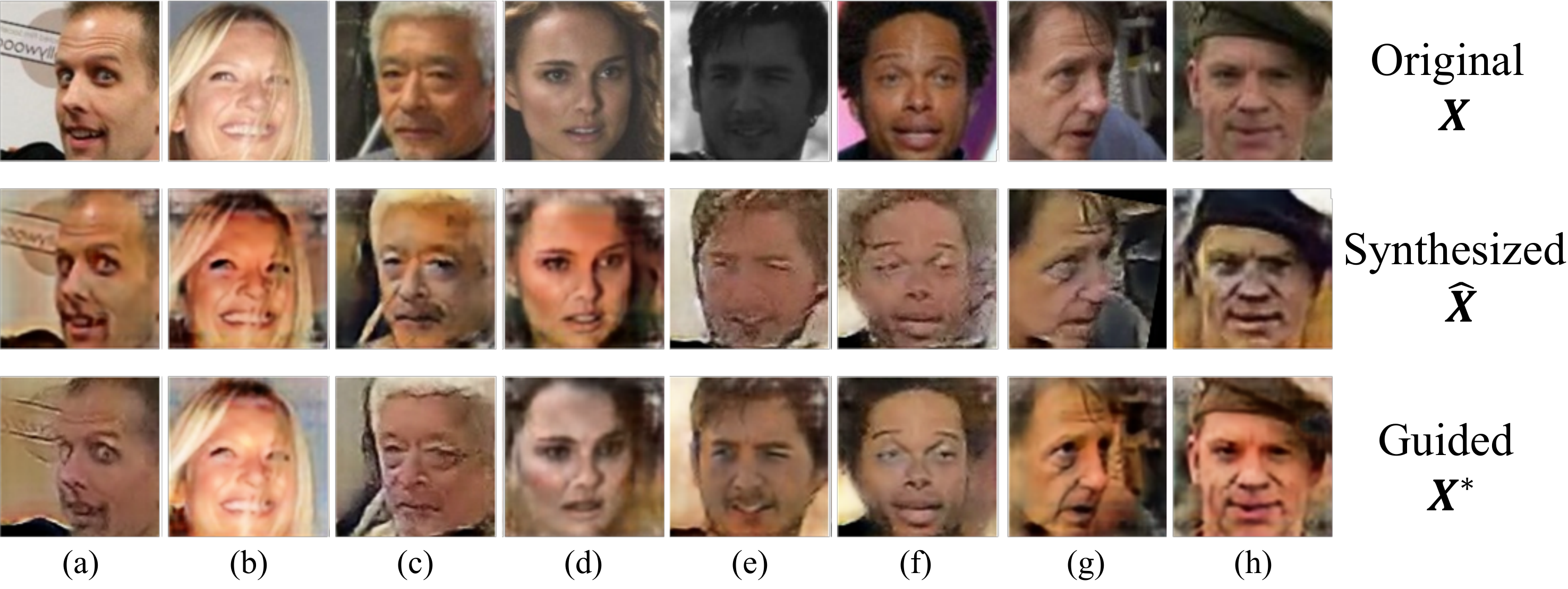}
\caption{Plot of mini-batch samples augmented with CFSM during training of the FR model. Top: Original images. Middle: Synthesized results before the feedback of the FR model. Bottom: Synthesized results after the feedback. The guide from the FR model can vary the images' style for increased difficulty (a-d), and preventing the images from identity lost (e-h).}
\label{fig:update_faces}
\end{wrapfigure}
where $\nabla_{\mathbf{}}\mathcal{L}_{cla}(\cdot,\cdot)$ denotes the gradient of $\mathcal{L}_{cla}(\cdot,\cdot)$ w.r.t.~$\mathbf{o}$, and $\text{sgn}(\cdot)$ is the sign function. 

Finally, based on the adversarial-based augmented face images, we further optimize the face embedding model $\mathcal{F}$ via the objective:
\begin{equation}
    \min_{\theta}\mathcal{L}_{cla}([\mathbf{X}^{*}, \mathbf{X}], l),
    \label{eqn:outer_min}
\end{equation}
where $\theta$ indicates the parameters of FR model $\mathcal{F}$ and $[\cdot]$ refers to concatenation in the batch dimension.
In other words, it encourages to search for the best perturbations in the learned style latent space in the direction of maximal difficulty for the FR model. Examples within a mini-batch are shown in Fig.~\ref{fig:update_faces}.

\Paragraph{Dataset Distribution Measure} 
As mentioned above, as a by-product of our learned face synthesis model, we obtain a target-specific linear subspace model, which can characterize the variations in the target dataset. Such learned linear subspace models allow us to quantify the distribution similarity among different datasets. For example, given two unlabeled datasets $A$ and $B$, we can learn the corresponding linear subspace models $\{\mathbf{U}_{A}, \boldsymbol\mu_{A}\}$ and $\{\mathbf{U}_{B}, \boldsymbol\mu_{B}\}$.
We define the distribution similarity between them as
\begin{equation}
    \mathcal{S}(A,B) =\frac{1}{q}\left(\sum_{i}^{q}S_{C}(\mathbf{u}_{A}^{i}+\boldsymbol\mu_{A}, \mathbf{u}_{B}^{i}+\boldsymbol\mu_{B})\right),
    \label{eqn:dist_simil}
\end{equation}
where $S_{C}(\cdot,\cdot)$ denotes the Cosine Similarity between the corresponding basis vectors in $\mathbf{U}_{A}$ and $\mathbf{U}_{B}$ respectively, and $q$ is the number of the basis vectors.  

Measuring the distance or similarity between datasets is a fundamental concept underlying research areas such as domain adaptation and transfer learning. 
However, the solution to this problem typically involves measuring feature distance with respect to a learned model, which may be susceptible to modal failure that the model may encounter in unseen domains. 
In this work, we provide an alternative solution via learned style bases vectors, that are directly optimized to capture the characteristics of the target dataset. We hope our method could provide new understandings and creative insights in measuring the dataset similarity.
For visualizations of $S$ among different datasets, please refer to Sec.~\ref{sec:exp_dist_simi}.

\subsection{Implementation Details}
All face images are aligned and resized into $112\times 112$ pixels.
The network architecture of the face synthesis model is given in the supplementary (\emph{\textbf{Supp}}). In the main experiments, we set $q{=}10$, $d{=}128$, $l_a{=}0$, $u_a{=}6$, $l_m{=}0.05$, $u_m{=}0.65$, $\lambda_{adv}{=}1$, $\lambda_{ort}{=}1$, $\lambda_{id}{=}8$, $\epsilon{=}0.314$. For more details, refer to Sec.~\ref{sec:exp} or \emph{\textbf{Supp}}.

\section{Experimental Results}\label{sec:exp}

\subsection{Comparison with SoTA FR methods}~\label{sec:comparison}
\Paragraph{Datasets}
Following the experimental setting of~\cite{shi2021boosting}, we use \textbf{MS-Celeb-1M}~\cite{guo2016ms} as our labeled training dataset. MS-Celeb-1M is a large-scale public face dataset with web-crawled celebrity photos. For a fair comparison, we use the cleaned MS$1$M-V$2$ ($3.9$M images of $85.7$K classes) from~\cite{shi2021boosting}.
For our target data, \textbf{WiderFace}~\cite{yang2016wider} is used. WiderFace is a dataset collected for face detection in challenging scenarios, with a diverse set of unconstrained variations. It is a suitable target dataset for training CFSM, as we aim to bridge the gap between the semi-constrained training faces and the faces in challenging testing scenarios. We follow~\cite{shi2021boosting} and use $70$K face images from WiderFace.
For evaluation, we test on four \textbf{unconstrained} face recognition benchmarks: IJB-B, IJB-C, IJB-S and TinyFace. These $4$ datasets represent real-world testing scenarios where faces are significantly different from the semi-constrained training dataset. 

  %
$\diamond$ \textbf{IJB-B}~\cite{whitelam2017iarpa} contains both high-quality celebrity photos collected in the wild and low-quality photos or video frames with large variations.  
  It consists of $1,845$ subjects with $21.8$K still images and $55$K frames from $7,011$ videos. 

$\diamond$ \textbf{IJB-C}~\cite{maze2018iarpa} is an extension of IJB-B, which includes about $3,500$ subjects with a total of $31,334$ images and $117,542$ unconstrained video frames.

$\diamond$ \textbf{IJB-S}~\cite{kalka2018ijb} is an extremely challenging benchmark where the images were collected in real-world surveillance environments. The dataset contains $202$ subjects with an average of $12$ videos per subject. Each subject also has $7$ high-quality enrollment photos under different poses. We test on three protocols, Surveillance-to-Still (\textbf{V2S}), Surveillance-to-Booking (\textbf{V2B}) and Surveillance-to-Surveillance (\textbf{V2V}). The first/second notation in the protocol refers to the probe/gallery image source. `Surveillance' (V) refers to the surveillance video, `still' (S) refers to the frontal high-quality enrollment image and `Booking' (B) refers to the $7$ high-quality enrollment images.

$\diamond$ \textbf{TinyFace}~\cite{cheng2018low} consists of $5,139$ labelled facial identities given by $169,403$ native low resolution face images, which is created to facilitate the investigation of unconstrained low-resolution face recognition.

\begin{table}[t]
\renewcommand\arraystretch{1.00}
  \caption{ Comparison with state-of-the-art methods on the IJB-B benchmark. \quad\quad\quad `*' denotes a subset of data selected by the authors.}
  \centering
  \resizebox{0.92\linewidth}{!}{
  \begin{tabular}{l |c |c || c |c |c |c |c}
    \hline
    \multirow{2}{*}{Method} & \multirow{2}{*}{ Train Data, \#labeled(+\#unlabeled)   }  & \multirow{2}{*}{Backbone}   & 
    \multicolumn{3}{c|}{Verification}   &    \multicolumn{2}{c}{Identification}    \\
    \cline{4-8}
    & & & $1e-5$ & $1e-4$ & $1e-3$ & Rank$1$ & Rank$5$ \\
   \hline
    VGGFace$2$~\cite{cao2018vggface2}  & VGGFace$2$, $3.3$M & SE-ResNet-$50$ & $70.50$ & $83.10$ & $90.80$ & $90.20$ & $94.6$ \\
    AFRN~\cite{kang2019attentional}  & VGGFace$2$-*, $3.1$M & ResNet-$101$  & $77.10$ & $88.50$ & $94.90$ & $\textbf{97.30}$ & $\textbf{97.60}$\\
    ArcFace~\cite{deng2019arcface}  & MS$1$MV$2$, $5.8$M & ResNet-$50$ & $84.28$ & $91.66$  & $94.81$  & $92.95$ & $95.60$ \\
    MagFace~\cite{meng2021magface}  & MS$1$MV$2$, $5.8$M &  ResNet-$50$ & $83.87$ & $91.47$ & $94.67$ & $-$ & $-$ \\
    \hline
    Shi~\emph{et al.}~\cite{shi2021boosting}  & cleaned MS$1$MV2, $3.9$M(+$70$K) & ResNet-$50$  & $88.19$ & $92.78$ & $95.86$ & $95.86$ & $96.72$\\
    
    \hline
    \textbf{ArcFace}  & cleaned MS$1$MV2, $3.9$M & ResNet-$50$ & $87.26$ & $94.01$ & $95.95$ & $94.61$ & $96.52$ \\
    \textbf{ArcFace+Ours}    & cleaned MS$1$MV2, $3.9$M(+$70$K) & ResNet-$50$ & $\textbf{90.95}$ & $\textbf{94.61}$ & $\textbf{96.21}$ & $94.96$ & $96.84$ \\
  \hline
  \end{tabular}
  }
  \label{tab:result_ijbb}
\end{table}

\begin{table}[t]

\renewcommand\arraystretch{1.00}
  \caption{ Comparison with state-of-the-art methods on the IJB-C benchmark.}
  \centering
  \resizebox{0.92\linewidth}{!}{
  \begin{tabular}{l |c |c || c |c |c |c |c}
    \hline
    \multirow{2}{*}{Method} & \multirow{2}{*}{ Train Data, \#labeled(+\#unlabeled)}  & \multirow{2}{*}{Backbone}   & 
    \multicolumn{3}{c|}{Verification}   &    \multicolumn{2}{c}{Identification}    \\
    \cline{4-8}
    & & & $1e-6$ & $1e-5$ & $1e-4$ & Rank$1$ & Rank$5$ \\
   \hline
    VGGFace$2$~\cite{cao2018vggface2}  & VGGFace$2$, $3.3$M & SE-ResNet-$50$ & - & $76.80$ & $86.20$ & $91.40$ & $95.10$ \\
    AFRN~\cite{kang2019attentional}  & VGGFace$2$-*, $3.1$M & ResNet-$101$ & - & $88.30$ & $93.00$ & $95.70$ & $\textbf{97.60}$ \\
    PFE~\cite{shi2019probabilistic}  & MS$1$M-$*$, $4.4$M & ResNet-$64$ & - & $89.64$ & $93.25$ & $95.49$ & $97.17$ \\
    DUL~\cite{chang2020data}  & MS$1$M-$*$, $3.6$M & ResNet-$64$ & - & $90.23$ & $94.20$ & $95.70$ & $97.60$ \\
    ArcFace~\cite{deng2019arcface}  & MS$1$MV$2$, $5.8$M & ResNet-$50$ & $80.52$ & $88.36$ & $92.52$ & $93.26$ & $95.33$ \\
    MagFace~\cite{meng2021magface}  & MS$1$MV$2$, $5.8$M &  ResNet-$50$ & $81.69$ & $88.95$ & $93.34$ & $-$ & $-$ \\
    \hline
    Shi~\emph{et al.}~\cite{shi2021boosting}  & cleaned MS$1$MV2, $3.9$M(+$70$K) & ResNet-$50$  & $87.92$ & $91.86$ & $94.66$ & $95.61$ & $97.13$\\
    
    \hline
    \textbf{ArcFace}  & cleaned MS$1$MV2, $3.9$M & ResNet-$50$ & $87.24$ & $93.32$ & $95.61$ & $95.89$ & $97.08$ \\
    
    \textbf{ArcFace+ours}    & cleaned MS$1$MV2, $3.9$M(+$70$K) & ResNet-$50$ & $\textbf{89.34}$ & $\textbf{94.06}$ & $\textbf{95.90}$ & $\textbf{96.31}$ & $97.48$ \\
  \hline
  \end{tabular}
  }
  \label{tab:result_ijbc}
\end{table}

\begin{table}[t]
 \newcommand{\tabincell}[2]{\begin{tabular}{@{}#1@{}}#2\end{tabular}} 
\renewcommand\arraystretch{1.00}
  \caption{ Comparison with state-of-the-art methods on three protocols of the IJB-S and TinyFace benchmark. The performance is reported in terms of rank retrieval (closed-set) and TAR@FAR (open-set). It is worth noting that MARN~\cite{gong2019low} is a multi-mode aggregation method and is fine-tuned on UMDFaceVideo~\cite{bansal2017s}, a video dataset.}
  \centering
  \resizebox{1\linewidth}{!}{
  \begin{tabular}{l | c | c ||  c c c  c |c c  c c |c c  c c||c c}
    \hline
    \multirow{2}{*}{Method } & \multirow{2}{*}{\tabincell{c}{\textbf{Labeled}\\Train Data}} & \multirow{2}{*}{Backbone}  & 
    \multicolumn{4}{c|}{IJB-S V2S}   &    \multicolumn{4}{c|}{IJB-S V2B}   &
    \multicolumn{4}{c||}{IJB-S V2V} &  \multicolumn{2}{c}{TinyFace}  \\
    \cline{4-7}
    \cline{8-11}
    \cline{12-15}
    \cline{16-17}
    & & &Rank$1$ & Rank$5$ & $1\%$ & $10\%$ 
    &Rank$1$ & Rank$5$ & $1\%$ & $10\%$
    &Rank$1$ & Rank$5$ & $1\%$ & $10\%$ 
    &Rank$1$ & Rank$5$\\
    \hline
    C-FAN~\cite{gong2019video} & MS$1$M-$*$ & ResNet-$64$ & $50.82$ & $61.16$  & $16.44$ & $24.19$ 
                 & $53.04$ & $62.67$  & $27.40$ & $29.70$
                 & $10.05$ & $17.55$  & $0.11$ & $0.68$
                 & $-$ & $-$ \\
    MARN~\cite{gong2019low} & MS$1$M-$*$  & ResNet-$64$ & $58.14$ & $64.11$ & $21.47$ & $-$ 
                 & $59.26$ & $65.93$ & $32.07$ & $-$
                 & $22.25$ & $34.16$ & $0.19$ & $-$
                 & $-$ & $-$ \\
    PFE~\cite{shi2019probabilistic} & MS$1$M-$*$ & ResNet-$64$  & $50.16$ & $58.33$ & $31.88$ & $35.33$ 
                 & $53.60$ & $61.75$ & $35.99$ & $39.82$
                 & $9.20$ & $20.82$  & $0.84$ & $2.83$
                 & $-$ & $-$ \\
    ArcFace~\cite{deng2019arcface} & MS$1$MV2 & ResNet-$50$ & $50.39$ & $60.42$ & $32.39$ & $42.99$ 
                    & $52.25$ & $61.19$ & $34.87$ & $43.50$
                    & $-$ & $-$ & $-$ & $-$
                    & $-$ & $-$ \\
    \hline
    Shi~\emph{et al.}~\cite{shi2021boosting} & MS$1$MV2-* & ResNet-$50$ & $59.29$ & $66.91$ & $39.92$ & $50.49$ 
                 & $60.58$ & $67.70$ & $32.39$ & $44.32$
                 & $17.35$ & $28.34$ & $1.16$ & $5.37$
                 & $-$ & $-$ \\\hline
    \textbf{ArcFace}~\cite{deng2019arcface} & MS$1$MV2-*  & ResNet-$50$   & $58.78$ & $66.40$ & $40.99$ & $50.45$ 
    & $60.66$ & $67.43$ & $43.12$ & $51.38$ 
    & $14.81$ & $26.72$ & $2.51$ & $5.72$ 
    & $62.21$ & $66.85$ \\
    
    \textbf{ArcFace+Ours*} & MS$1$MV2-* & ResNet-$50$     & $61.69$ & $68.33$ & $43.99$ & $53.34$ 
    & $62.20$ & $69.50$ & $44.38$ & $53.49$ 
    & $18.14$ & $31.34$ & $2.09$ & $4.51$ 
    & $62.39$ & $67.36$ \\
    
    \textbf{ArcFace+Ours} & MS$1$MV2-* & ResNet-$50$  & $\textbf{63.86}$ & $\textbf{69.95}$ & $\textbf{47.86}$ & $\textbf{56.44}$ 
    & $\textbf{65.95}$ & $\textbf{71.16}$ & $\textbf{47.28}$ & $\textbf{57.24}$ 
    & $\textbf{21.38}$ & $\textbf{35.11}$ & $\textbf{2.96}$ & $\textbf{7.41}$ 
    & $\textbf{63.01}$ & $\textbf{68.21}$ \\ \hline
    
        \textbf{AdaFace}~\cite{kim2022adaface}  & WebFace12M & IResNet-$100$   & $71.35$ & $76.24$ & $59.40$ & $\textbf{66.34}$ 
    & $71.93$ & $76.56$ & $59.37$ & $\textbf{66.68}$ 
    & $36.71$ & $50.03$ & $4.62$ & $11.84$ 
    & $72.29$ & $74.97$ \\
    
        \textbf{AdaFace+Ours}  & WebFace12M & IResNet-$100$   & $\textbf{72.54}$ & $\textbf{77.59}$ & $\textbf{60.94}$ & $66.02$ 
    & $\textbf{72.65}$ & $\textbf{78.18}$ & $\textbf{60.26}$ & $65.88$ 
    & $\textbf{39.14}$ & $\textbf{50.91}$ & $\textbf{5.05}$ & $\textbf{13.17}$ 
    & $\textbf{73.87}$ & $\textbf{76.77}$ \\
    
    \hline
  \end{tabular}
  }
  \vspace{0mm}
  \label{tab:result_ijbs}
\end{table}

\Paragraph{Experiment Setting}
We first train CFSM with $\sim 10\%$ of MS-Celeb-1M training data ($n=0.4$M) as the source domain, and WiderFace as the target domain ($m=70$K). The model is trained for $125,000$ steps with a batch size of $32$. Adam optimizer is used with $\beta_{1}=0.5$ and $\beta_{2}=0.99$ at a learning rate of $1e-4$.

For the FR model training, we adopt ResNet-$50$ as modified in~\cite{deng2019arcface} as the backbone and use ArcFace loss function~\cite{deng2019arcface} for training. 
We also train a model without using CFSM (\emph{i.e.,} replication of ArcFace) for comparison, denoted as \textbf{ArcFace}.
The efficacy of our method (\textbf{ArcFace+Ours}) is validated by training an FR model with the guided face synthesis as the auxiliary data augmentation during training according to Eq.~\ref{eqn:outer_min}. 
%


\Paragraph{Results.}
Tables~\ref{tab:result_ijbb} and~\ref{tab:result_ijbc} respectively show the face verification and identification results on IJB-B and IJB-C datasets. 
Our approach achieves SoTA performance on most of the protocols.
For IJB-B, performance increase from using CFSM (\textbf{ArcFace+Ours}) is $3.69\%$ for TAR@FAR=$1e-5$, and $2.10\%$ for TAR@FAR=$1e-6$ on IJB-C. Since both IJB-B and IJB-C are a mixture of high quality images and low quality videos, the performance gains with the augmented data indicate that our model can generalize to both high and low quality scenarios.
In Tab.~\ref{tab:result_ijbs}, we show the comparisons on IJB-S and TinyFace. With our CFSM (\textbf{ArcFace+Ours}), ArcFace model outperforms all the baselines in both face identification and verification tasks, and achieves a new SoTA performance.

\begin{figure}[t]
\footnotesize
  \resizebox{1\linewidth}{!}{
\begin{tabular}{ c  c  c }

\raisebox{-.5\height}{\includegraphics[scale=0.3]{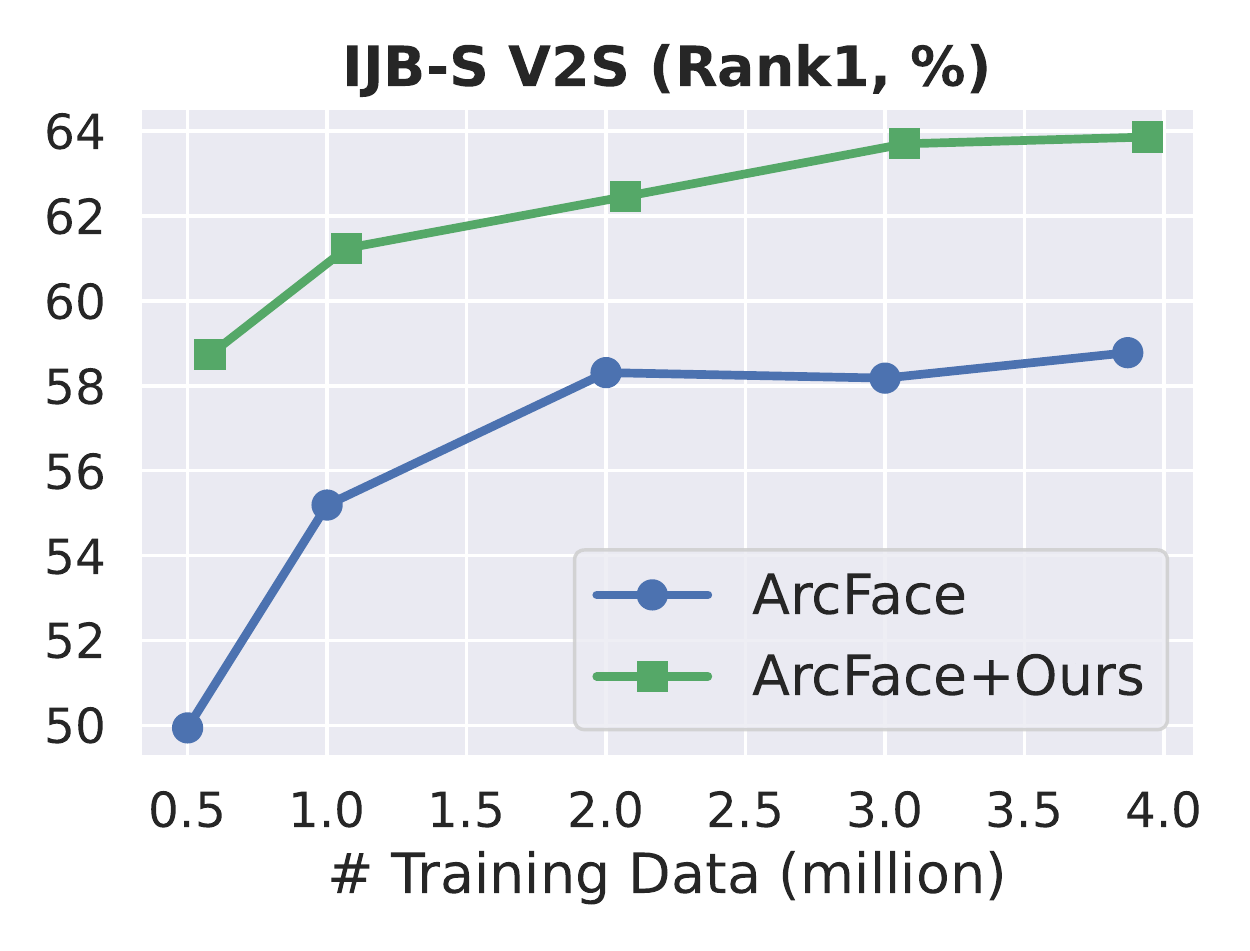}} &
\raisebox{-.5\height}{\includegraphics[scale=0.3]{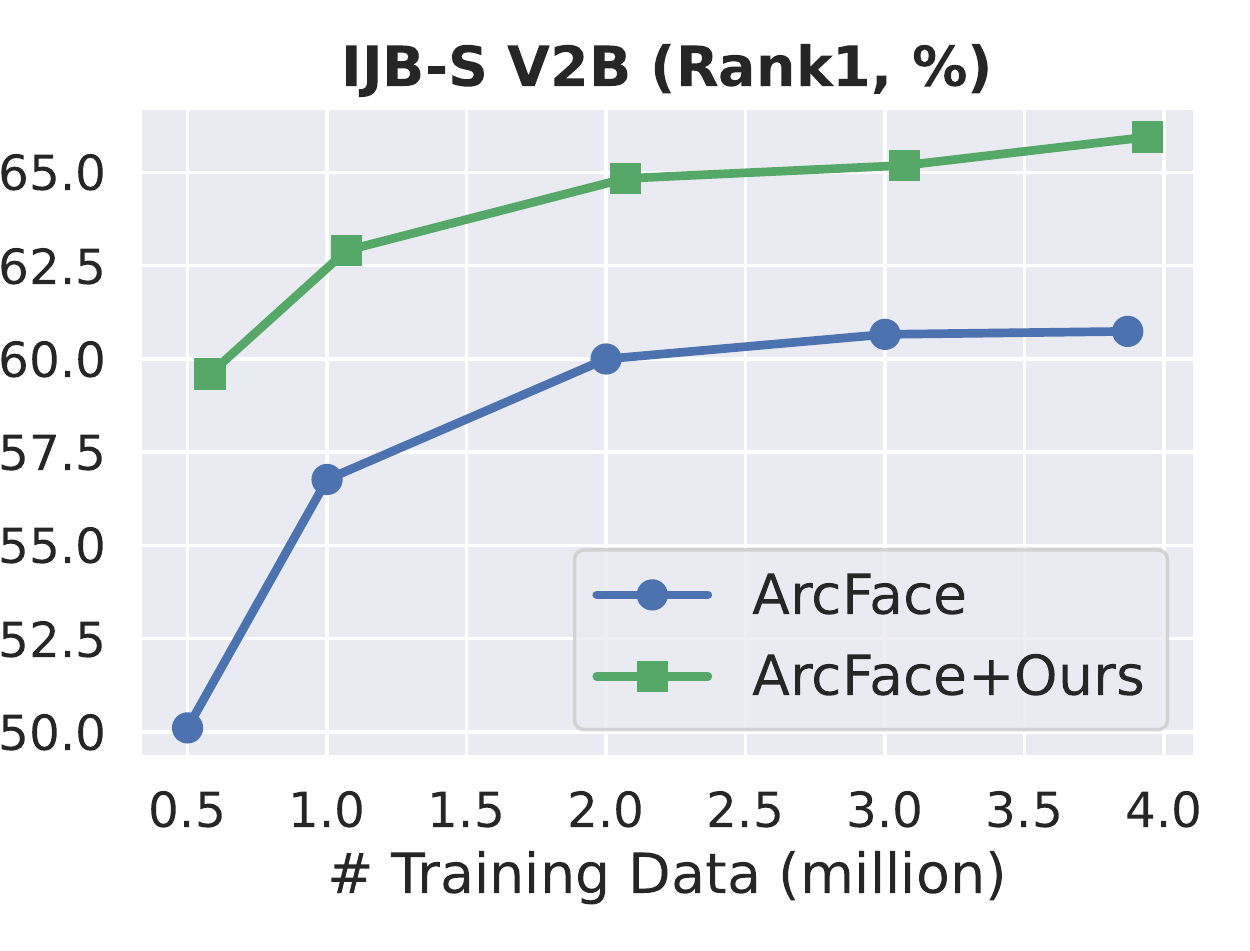}} &
\raisebox{-.5\height}{\includegraphics[scale=0.3]{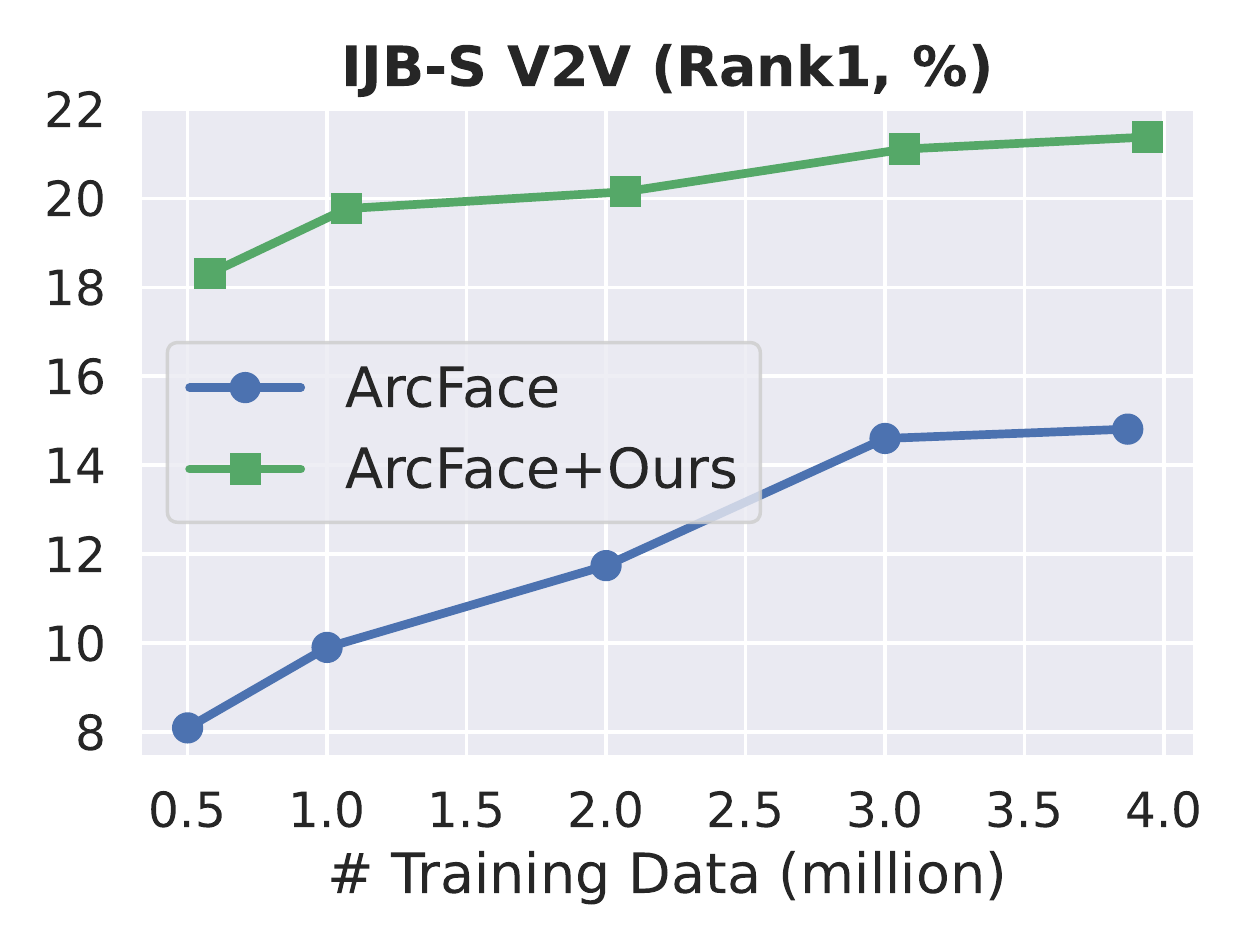}} \\
  \vspace{-5mm}\\ 
\raisebox{-.5\height}{\includegraphics[scale=0.3]{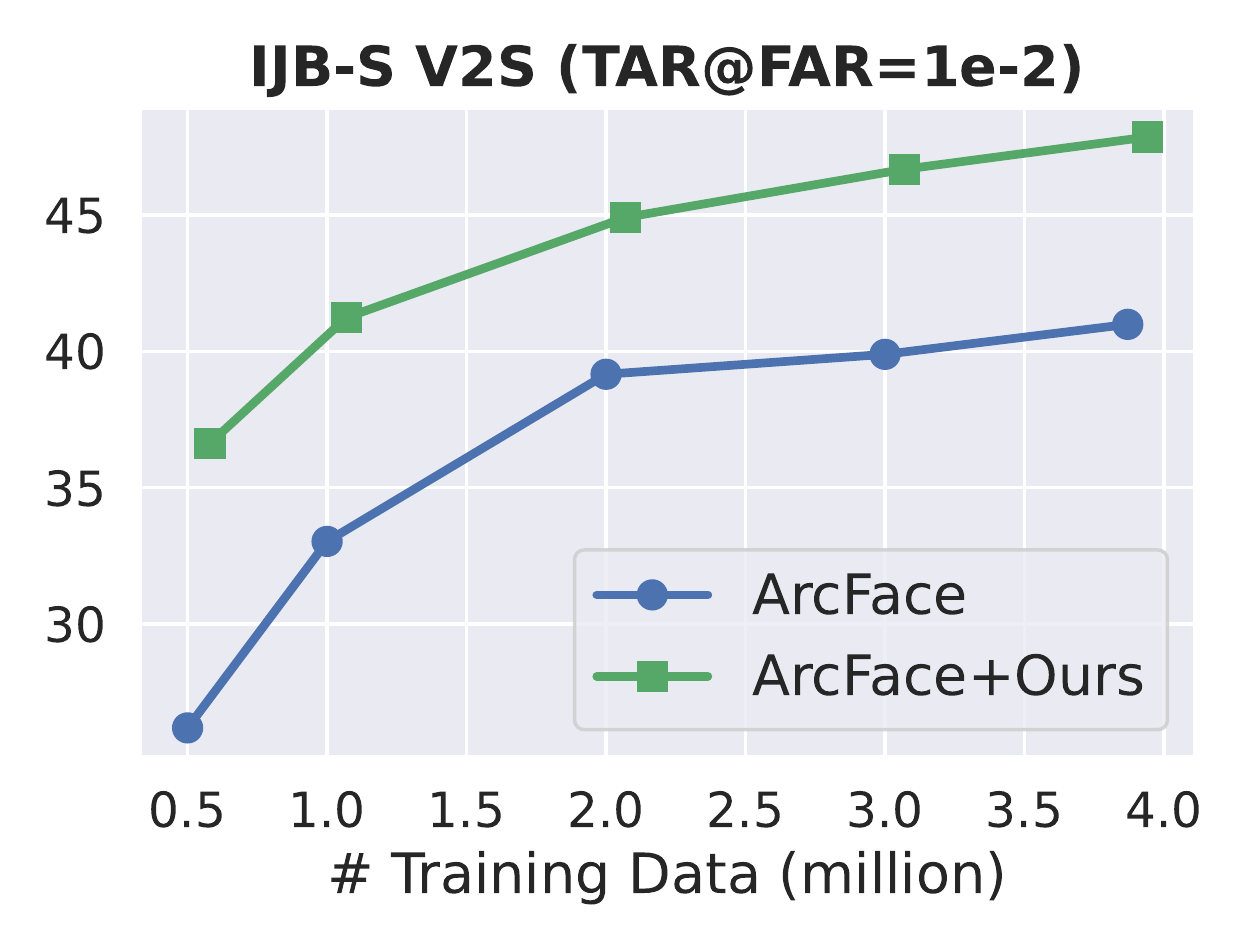}} &
\raisebox{-.5\height}{\includegraphics[scale=0.3]{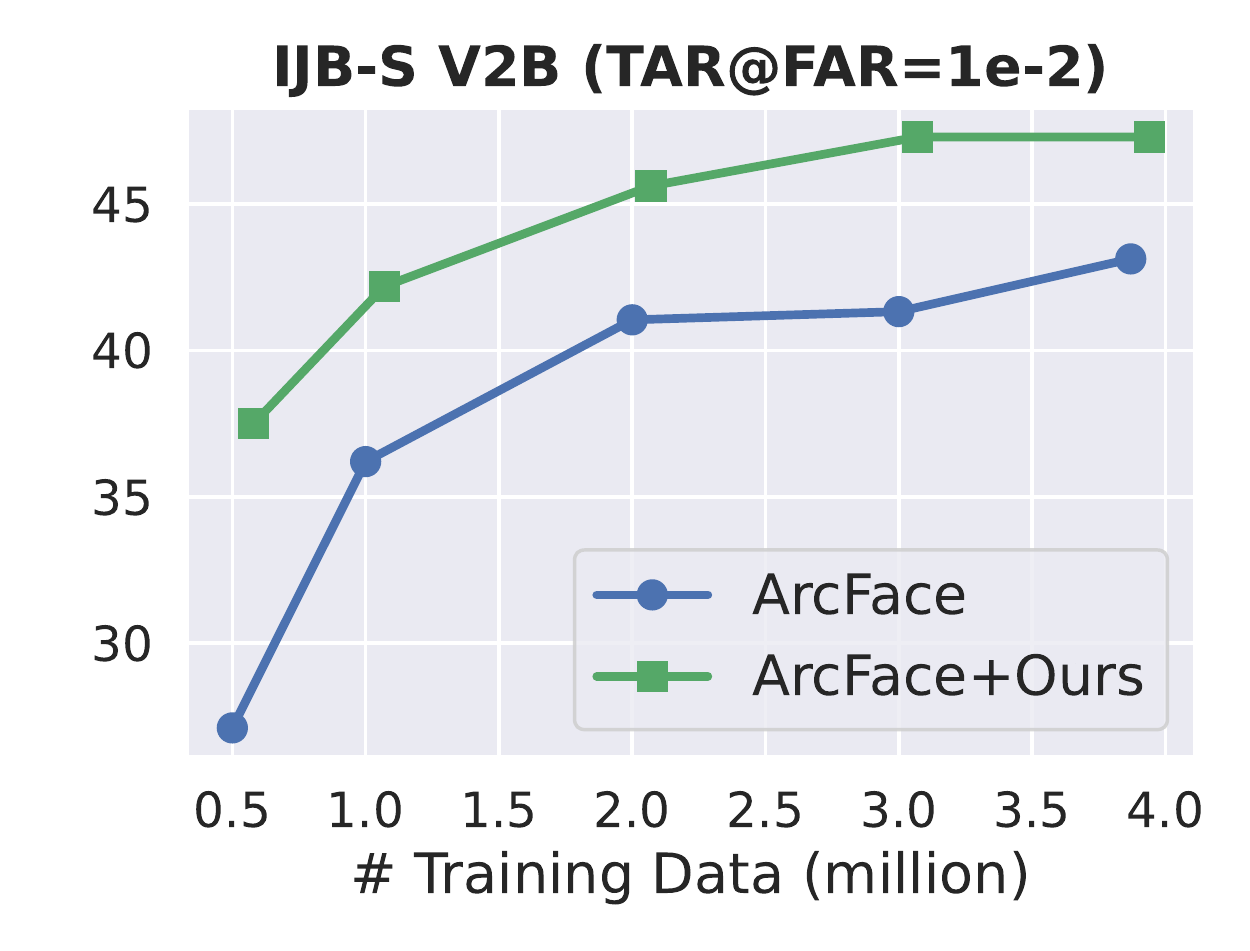}} &
\raisebox{-.5\height}{\includegraphics[scale=0.3]{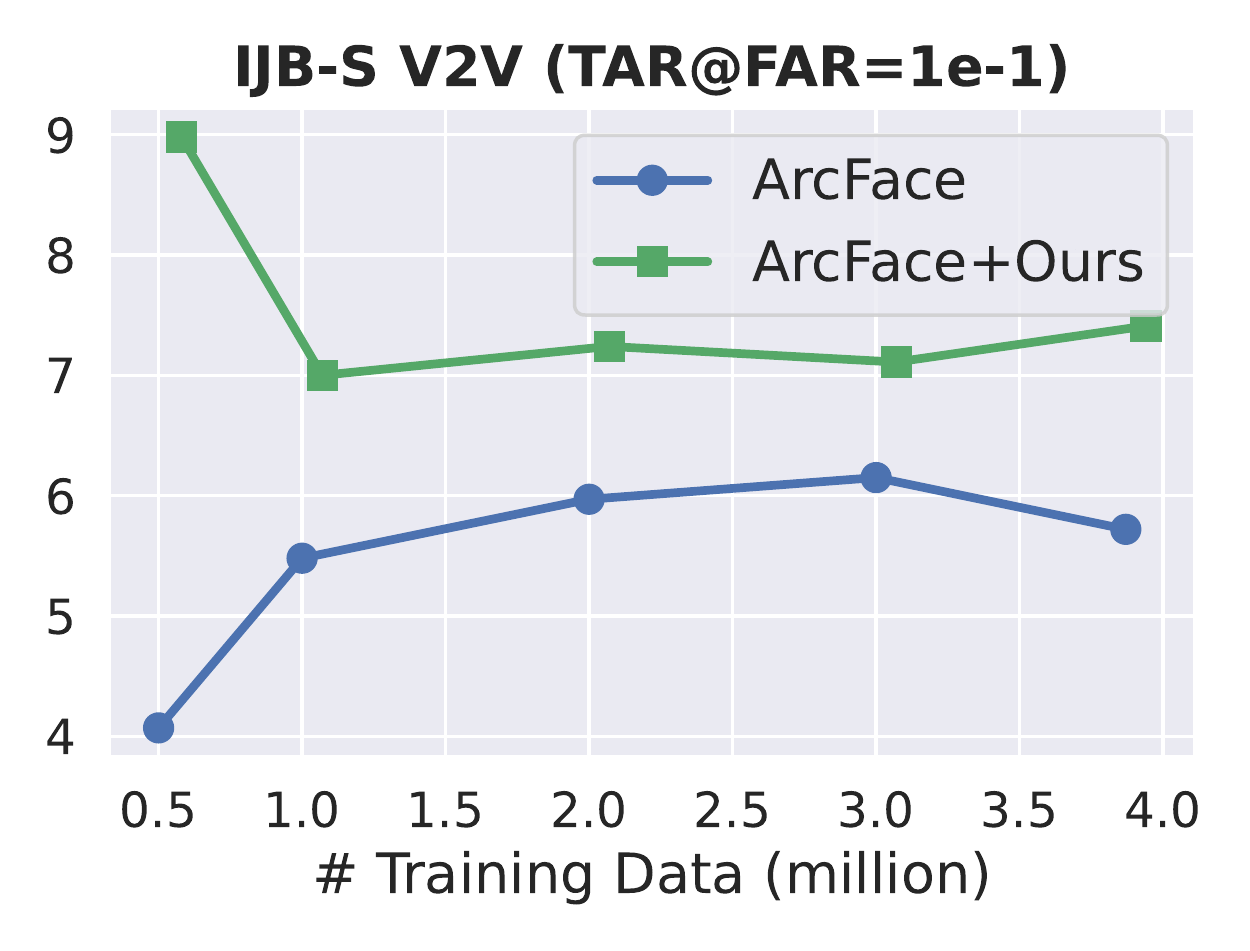}}
\end{tabular}
}
\vspace{0mm}
\caption{ Comparison on three IJB-S protocols 
with varied number of training data on the $x$-axis (maximum is 3.9M). The plots shows that using guided CFSM as an augmentation (\textcolor{mygreen}{\textbf{ArcFace+Ours}}) can lead to higher performance in all settings. Note that CFSM trained on $70$K unlabeled data is more useful than $3$M original data as shown by the higher \textbf{V2V} performance of \textbf{Ours} with $0.5$M than Baseline $3.9$M. } 
\label{fig:diff_training}
\vspace{-2mm}
\end{figure}


\subsection{Ablation and Analysis}
In this experiment, 
we compare the face verification and identification performance on the \emph{most challenging} IJB-S and TinyFace datasets.  

\Paragraph{Large-scale Training Data vs.~Augmentation.} To further validate the applicability of our CFSM as an augmentation in different training settings, we adopt IResNet-$100$~\cite{deng2019arcface} as the FR model backbone and utilize SoTA AdaFace loss function~\cite{kim2022adaface} and large-scale WebFace12M~\cite{zhu2021webface260m} dataset for training. As compared in Tab.~\ref{tab:result_ijbs}, our model still improves unconstrained face recognition accuracies by a promising margin (Rank1: $+2.43\%$ on the IJB-S V2V protocol and $+1.58\%$ on TinyFace) on a large-scale training dataset (WebFace12M).

\Paragraph{Effect of Guidance in CFSM.}
To validate the effectiveness of the proposed \emph{controllable} and \emph{guided} face synthesis model in face recognition, we train a FR model with CFSM as augmentation but with random style coefficients (\textbf{Ours*}), and compare with guided CFSM  (\textbf{Ours}). Tab.~\ref{tab:result_ijbs} shows that the synthesis with random coefficients does not bring significant benefit to unconstrained IJB-S dataset performance. However, when samples are generated with guided CFSM, the trained FR model performs much better. Fig.~\ref{fig:update_faces} shows the effect of guidance in the training images. For low quality images, too much degradation leads to images with altered identity. Fig.~\ref{fig:update_faces} shows that guided CFSM avoids synthesizing bad quality images that are unidentifiable.

\begin{figure}[t]

\centering    
\includegraphics[width=10.0cm]{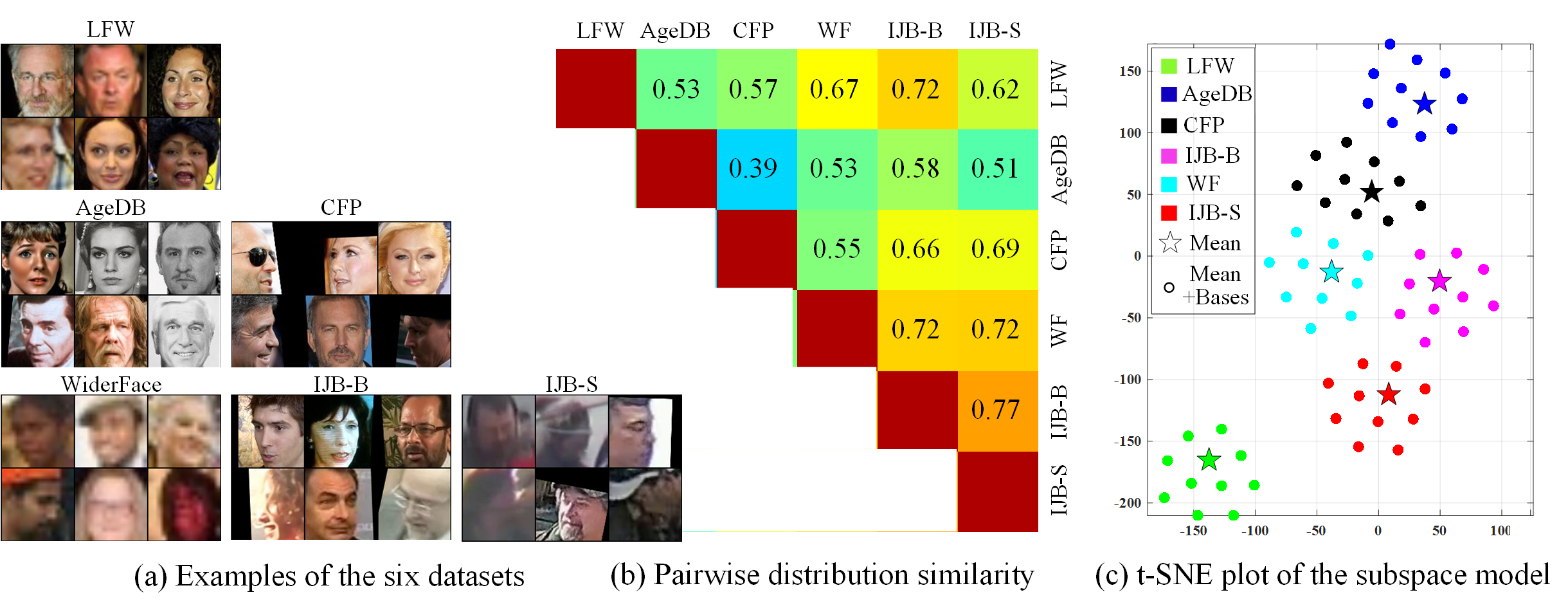}
\vspace{-2mm}
\caption{ \textbf{(a)} Few examples from each dataset. Note differences in style. For example, AgeDB contains old grayscale photos, WiderFace has mostly low resolution faces, and IJB-S includes extreme unconstrained attributes (\emph{i.e.,} noise, motion blur or turbulence effect). \textbf{(b)} shows the pairwise distribution similarity scores among datasets that are calculated using the learned subspace via Eq.~\ref{eqn:dist_simil}. Note both IJB-B and WiderFace have high similarity scores with IJB-S. \textbf{(c)} The t-SNE plot of the learned $[\mathbf{u}_1,...,\mathbf{u}_{10}]$ and mean style $\mathbf{\mu}$. The dots represent $\mathbf{u}_{i}+\mathbf{\mu}$ and the stars denote $\mathbf{\mu}$.} 
\vspace{-2mm}
\label{fig:dataset}
\end{figure}
 

\begin{figure}[t]
\centering
\includegraphics[width=11cm]{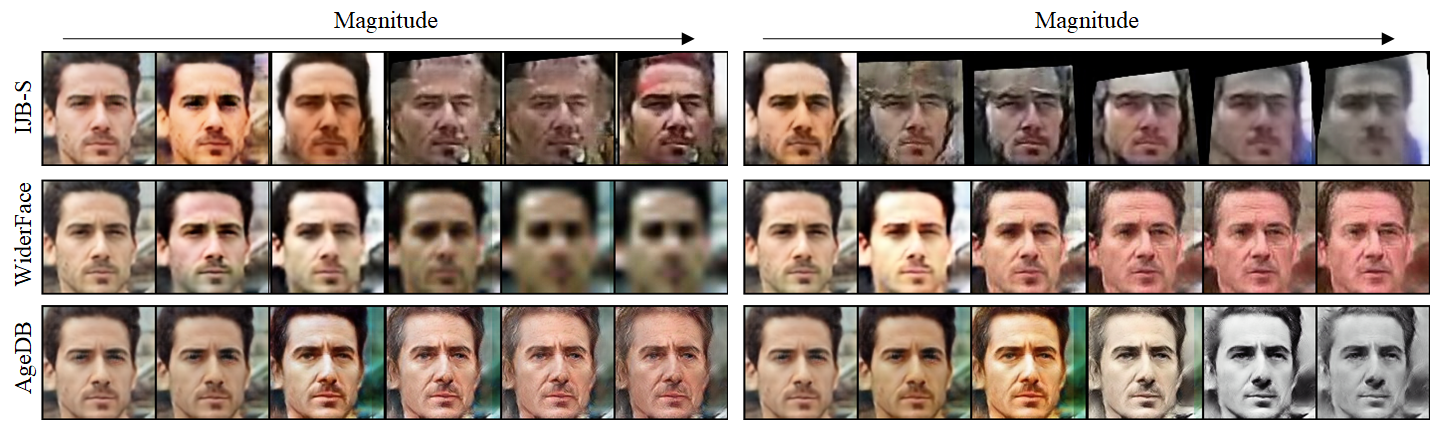}
\caption{ \textbf{Interpretable magnitude of the style coefficients}. Given an input image, we randomly sample two sets of style coefficients $\mathbf{o_{1}}$ (left) and $\mathbf{o_{2}}$ (right) for all $3$ models (respectively trained with the IJB-S, WiderFace and AgeDB datasets as the target data). We dynamically adjust the magnitude of these two coefficients by $0.5a$, $a$, $1.5a$, $2a$, $3a$, $4a$, where $a=\frac{\mathbf{o}}{||\mathbf{o}||}$.
As can be seen, our model indeed realizes the goal of changing the degree of style synthesis with the coefficient magnitude.
}
\label{fig:mag_examples}
\end{figure}

\Paragraph{Effect of the Number of Labeled Training Data.}
To validate the effect of the number of labeled training data, we train a series of models by adjusting the number of labeled training samples from $0.5$M to $3.9$M and report results both on \textbf{ArcFace} and \textbf{ArcFace+Ours} settings. Fig.~\ref{fig:diff_training} shows the performance on various IJB-S protocols. For the full data usage setting, our model trained with guided CFSM as an augmentation outperforms the baseline by a large margin.
Also note that the proposed method trained with $1/8$th ($0.5$M) labeled data still achieves comparable performance, or even better than the baseline with $3.9$M labeled data on \textbf{V2V} protocol. This is due to CFSM generating target data-specific augmentations, thus demonstrating the value of our controllable and guided face synthesis, which can significantly boost unconstrained FR performance.

\subsection{Analysis and Visualizations of the Face Synthesis model}\label{sec:exp_dist_simi}
In this experiment, we quantitatively evaluate the distributional similarity between datasets based on the learned linear subspace model for face synthesis. To this end, we choose $6$ face datasets that are publicly available and popular for face recognition testing. These datasets are LFW~\cite{huang2008labeled}, AgeDB-$30$~\cite{moschoglou2017agedb}, CFP-FP~\cite{sengupta2016frontal}, IJB-B~\cite{whitelam2017iarpa}, WiderFace (WF)~\cite{yang2016wider} and IJB-S~\cite{kalka2018ijb}. 
Figure~\ref{fig:dataset}\textcolor{red}{(a)} shows examples from these $6$ datasets. Each dataset has its own style. For example, CFP-FP includes profile faces, WiderFace has mostly low resolution faces, and IJB-S contains extreme unconstrained attributes. 
During training, for each dataset, we randomly select $12$K images as our target data to train the synthesis model. For the source data, we use the same subset of MS-Celeb-$1$M as in Sec.~\ref{sec:comparison}.

\Paragraph{Distribution Similarity}
Based on the learned dataset-specific linear subspace model, we calculate the pairwise distribution similarity score via Eqn.~\ref{eqn:dist_simil}. As shown in Fig.~\ref{fig:dataset}\textcolor{red}{(b)}, the score reflects the style correlation between datasets. For instance, strong correlations among IJB-B, IJB-S and WiderFace (WF) are observed. We further visualize the learned basis vectors $[\mathbf{u}_{1},...,\mathbf{u}_{q}]$ and the mean style $\mu$ in Fig.~\ref{fig:dataset}\textcolor{red}{(c)}. The basis vectors are well clustered and the discriminative grouping indicates the correlation between dataset-specific models.

\Paragraph{Visualizations of Style Latent Spaces}
Fig.~\ref{fig:mag_examples} shows face images generated by the learned CFSM. It can be seen that when the magnitude increases, the corresponding synthesized faces reveal more dataset-specific style variations. This implies the magnitude of the style code is a good  indicator of image quality.

\begin{figure}[t]
\footnotesize
\resizebox{1\linewidth}{!}{
\begin{tabular}{ c  c c }
   { LFW} & { AgeDB} & { CFP}\\
\raisebox{-.5\height}{\includegraphics[scale=0.28]{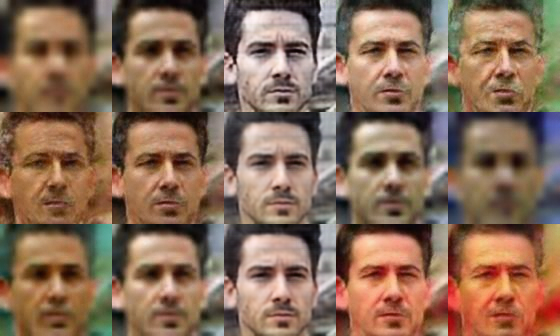}}  & \raisebox{-.5\height}{\includegraphics[scale=0.28]{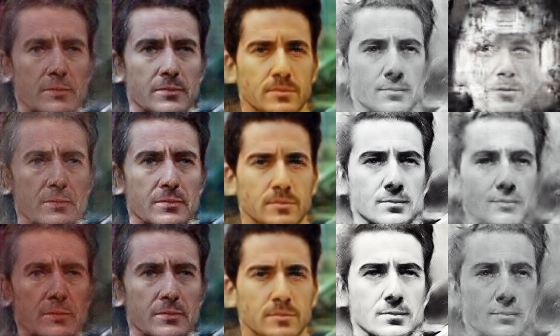}} & \raisebox{-.5\height}{\includegraphics[scale=0.28]{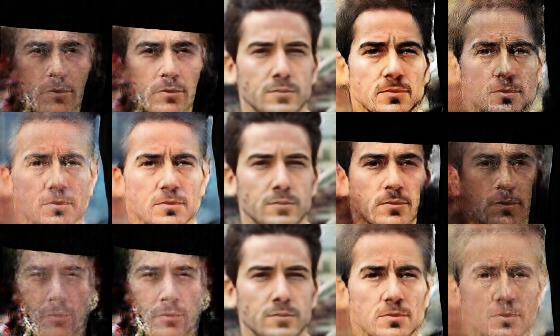}}  \\
\vspace{-2mm} \\
   { IJB-B} & { WiderFace} & { IJB-S}\\
\raisebox{-.5\height}{\includegraphics[scale=0.28]{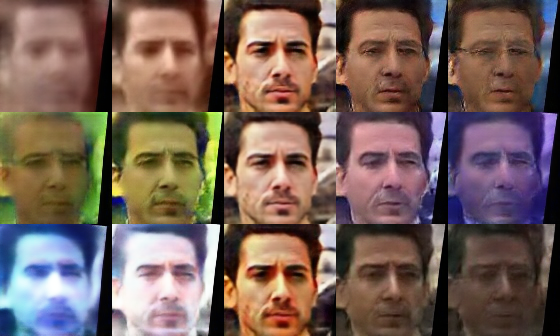}}  & \raisebox{-.5\height}{\includegraphics[scale=0.28]{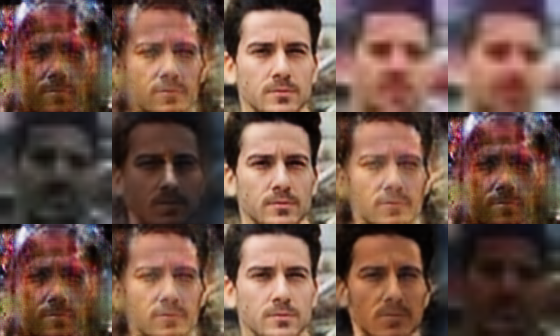}} & \raisebox{-.5\height}{\includegraphics[scale=0.28]{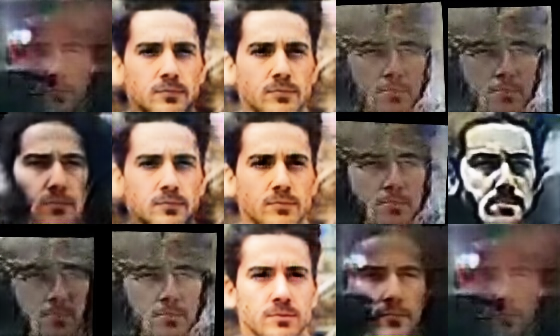}}  \\
\end{tabular}
}
\caption{
Given a single input image, we visualize the synthesized images by traversing along with the learned orthonormal basis $\mathbf{U}$ in the $6$ dataset-specific models. 
For each dataset, Rows $1{-}3$ illustrate the first $3$ basis vectors traversed. 
Columns $1{-}5$ show the directions which are scaled to emphasize their effect, {\it i.e.}, only one element of style coefficient $\mathbf{o}$ varies from $-3\sigma$ to $3\sigma$ while other $q\!-\!1$ elements remain $0$. }
\label{fig:basis_examples}
\end{figure}

We also visualize the learned $\mathbf{U}$ of $6$ models in Fig.~\ref{fig:basis_examples}. As we move along a basis vector of the learned subspace, the synthesized images change their style in dataset-specific ways. For instance, with target as WiderFace or IJB-S, synthesized images show various low quality styles such as blurring or turbulence effect. CFP dataset contains cropped images, and the ``crop'' style manifests in certain directions.  Also, we can observe the learned $\mathbf{U}$ are different among datasets, which further verifies that our learned linear subspace model in CFSM is able to capture the variations in target datasets.


\section{Conclusions}\label{sec:con}
We answer the fundamental question of ``\textit{How can image synthesis benefit the end goal of improving the recognition task?}'' 
Our controllable face synthesis model (CFSM) with adversarial feedback of FR model shows the merit of task-oriented image manipulation, evidenced by significant performance increases in unconstrained face datasets (IJB-B, IJB-C, IJB-S and TinyFace). 
In a broader context, it shows that adversarial manipulation could go beyond being an attacker, and serve to increase recognition accuracies in vision tasks.
Meanwhile, we define a dataset similarity metric based on the learned style bases, which capture the style differences in a label or predictor agnostic way.
We believe that our research has presented the power of a controllable and guided face synthesis model for unconstrained FR and provides an understanding of dataset differences.

\Paragraph{Acknowledgments.} This research is based upon work supported in part by the Office of the Director of National Intelligence (ODNI), Intelligence Advanced Research Projects Activity (IARPA), via 2022-21102100004. The views and conclusions contained herein are those of
the authors and should not be interpreted as necessarily representing the official policies, either expressed or implied, of ODNI, IARPA, or the U.S. Government. The U.S. Government is authorized to reproduce and distribute reprints for governmental purposes notwithstanding any copyright annotation therein.


\newpage
\section*{Supplementary}
\renewcommand{\thesection}{\Alph{section}}
\setcounter{section}{0}

In this supplementary material, we provide:

$\diamond$ Additional implementation details including the network structure of the face synthesis model and the training process.

$\diamond$ Additional ablation studies, including the effects of different target datasets, the dimensionality of the style coefficient, the perturbation budget, and the ratio of real and synthetic images in each mini-batch.

$\diamond$ Additional visualizations of the adversarial training process. 

\section{Additional Implementation Details}
\begin{figure}[t]
\centering
\includegraphics[width=12.0cm]{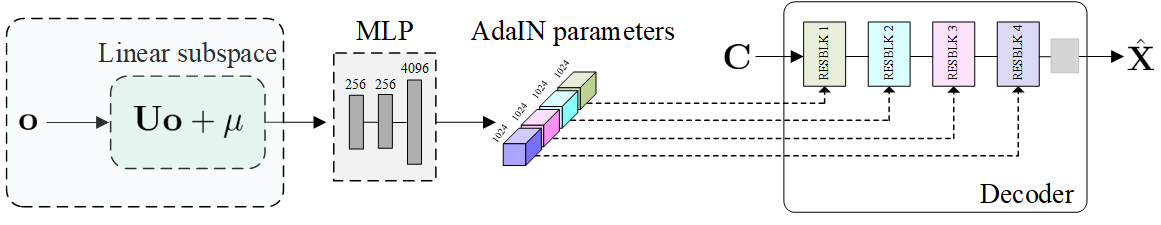}
\vspace{0mm}
\caption{\small Additional illustration of the decoder network structure. The parameters of Adaptive Instance Normalization (AdaIN) in residual blocks are dynamically generated by a multiplayer perceptron (MLP) from the linear subspace model.}
\label{fig:decoder}
\vspace{-2mm}
\end{figure}

\Paragraph{Network Structure.} The network architecture of the generator (including the encoder $E$ and decoder $G$) used in our face synthesis module is illustrated in Tab.~\ref{tab:generator}. We apply Instance Normalization~\cite{ulyanov2017improved} to the encoder and Adaptive Instance Normalization~\cite{huang2017arbitrary} to RESBLOKs (the residual basic block) of the decoder.
The encoder takes an image $\mathbf{X}$ with the resolution of $112\times112$ as input, and outputs its content feature $\mathbf{C}\in\mathbb{R}^{256\times28\times28}$. The input and output to the decoder are $\mathbf{C}$ and the synthesized image $\hat{\mathbf{X}}$, respectively. Additionally, as shown in Fig.~\ref{fig:decoder}, the parameters of the Adaptive Instance Normalization (AdaIN) layer in residual blocks are dynamically generated by a multiplayer perceptron (MLP) from the linear subspace model.
Following~\cite{wang2018high}, we employ multi-scale discriminators with $3$ scales as our discriminator $D$.

\Paragraph{Training Process.} We summarize the training process in Tab.~\ref{tab:train_process}. In Stage 1, we train our controllable face synthesis module with the identity consistency loss and the adversarial objective. In Stage 2, based on the pre-trained and fixed face synthesis model, we introduce an adversarial regularization strategy to guide the data augmentation process and train the face feature extractor $\mathcal{F}$.

Specifically, in the adversarial FR model training, given $B$ face images $\{\mathbf{X}\}_{i=1}^{B}$ in a mini-batch, our synthesis model (CFSM) is utilized to produces their synthesized version $\hat{\mathbf{X}}$ with initial random style coefficients $\{\mathbf{o}\}_{i=1}^{B}$. Based on the Eqn. 7 and 8 (main paper), we obtain the updated style coefficients $\{\mathbf{o}^{*}\}_{i=1}^{B}$ with perturbations. We then generate the perturbed images $\{\mathbf{X}^{*}\}_{i=1}^{B}$ with CFSM. Finally, we randomly select half of $\{\mathbf{X}\}_{i=1}^{B}$ and half of $\{\mathbf{X}^{*}\}_{i=1}^{B}$ to form a new training batch for the FR model training. Note that, every epoch of the FR model training we will randomly initialize different style coefficients, even for the same training samples.

\begin{table}[t]
\vspace{0mm}
\renewcommand\arraystretch{1.3}
  \caption{\small Network architectures of the generator of face synthesis module. RESBLK denotes the residual basic block. [Keys: N=Neurons, K=Kernel size, S=Stride, B=Batch size]. }
  \centering
  \vspace{0mm}
  \resizebox{0.8\linewidth}{!}{
  \begin{tabular}{c ||  c  ||c  }
    \hline
    Layer & Encoder (E) & Decoder (G) \\
    \hline
    1 & CONV-(N64,K7,S1), ReLU   &  RESBLK-(N256,K3,S1) \\
    2 & CONV-(N128,K4,S2), ReLU  &  RESBLK-(N256,K3,S1)  \\
    3 & CONV-(N256,K4,S2), ReLU  &  RESBLK-(N256,K3,S1)  \\
    4 & RESBLK-(N256,K3,S1) & RESBLK-(N256,K3,S1)   \\
    5 & RESBLK-(N256,K3,S1) & CONV-(N128,K5,S1), ReLU  \\
    6 & RESBLK-(N256,K3,S1) & CONV-(N64,K5,S1), ReLU  \\
    7 & RESBLK-(N256,K3,S1) & CONV-(N3,K7,S1), TanH  \\ \hline
    Output & $\mathbf{C}\in\mathbb{R}^{B\times256\times28\times28}$   & $\hat{\mathbf{X}}\in\mathbb{R}^{B\times3\times W\times H}$   \\ \hline

  \end{tabular}
  }
  \label{tab:generator}
  \vspace{0mm}
\end{table}

\begin{table}[t]
\vspace{0mm}
\renewcommand\arraystretch{1.3}
  \caption{\small Stages of the training process.}
  \centering
  \vspace{0mm}
  \resizebox{0.6\linewidth}{!}{
  \begin{tabular}{c |  c  |c  }
    \hline
      & Network or parameters & Loss \\
    \hline \hline
    Stage 1 & $E$, $G$, $D$, MLP, $\mathbf{U}$, $\mu$  & $\mathcal{L}_{ort}$, $\mathcal{L}_{adv}$, $\mathcal{L}_{D}$, $\mathcal{L}_{id}$ \\
    \hline
    Stage 2 & $\mathcal{F}$, $\delta^{*}$ & $\mathcal{L}_{cla}$ \\
    \hline
  \end{tabular}
  }
  \label{tab:train_process}
  \vspace{-2mm}
\end{table}


\section{Additional Ablation Studies}
\begin{figure}[t]
\vspace{0mm}
\footnotesize
  \resizebox{1\linewidth}{!}{
\begin{tabular}{ c  c  c }

\raisebox{-.5\height}{\includegraphics[scale=0.3]{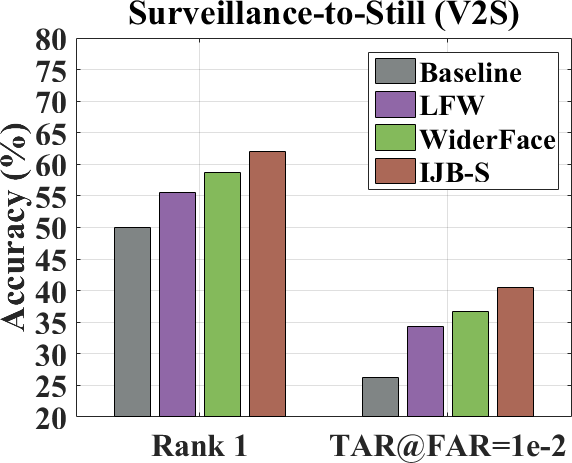}} &
\raisebox{-.5\height}{\includegraphics[scale=0.3]{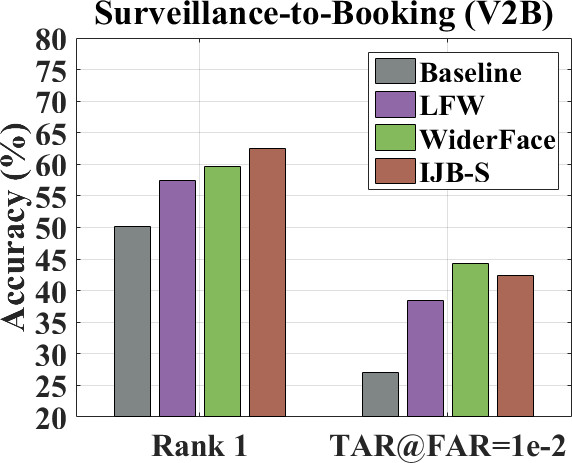}} &
\raisebox{-.5\height}{\includegraphics[scale=0.3]{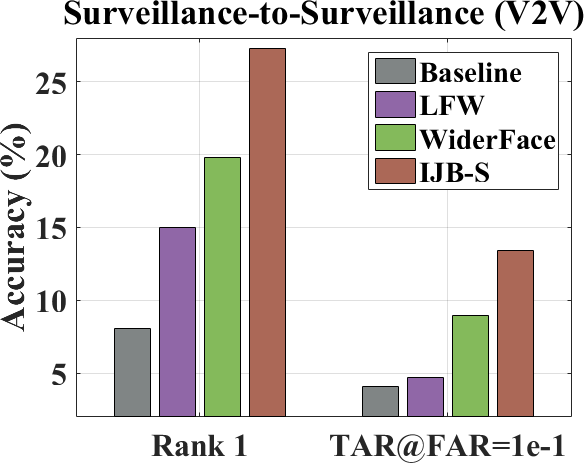}} \\
  \vspace{-4mm}\\ 
  (a)& (b) & (c) \\
\end{tabular}
}
\caption{\small Evaluation results on IJB-S with different target datasets. \textbf{Baseline} refers to the performance of the FR model trained on $0.5$ million labeled samples (a subset of MS-Celeb-1M) without using the proposed face synthesis model. In this experiment, other $3$ FR models are trained on the $0.5$ million labeled samples with the proposed face synthesis models, which are trained with additional $12$K unlabeled samples (from LFW, WiderFace or IJB-S, respectively).} 
\label{fig:diff_datasets}
\end{figure}

\Paragraph{Effect of Different Target Datasets.} 
To study how the choice of target dataset in face synthesis model training would affect the face recognition performance, we choose two other datasets, LFW~\cite{huang2008labeled} and IJB-S~\cite{{kalka2018ijb}} to train the face synthesis models and apply them for the FR model training. During training, for each dataset, we randomly select \emph{unlabeled} $12$K face images as the target data to train the face synthesis model. For efficiency, we train the FR models with $0.5$M labeled training samples from the MS-Celeb-1M dataset. The diversity of the three face datasets can be ranked as IJB-S $>$ WiderFace $>$ LFW. We show the comparisons on IJB-S protocols in Fig.~\ref{fig:diff_datasets}, which shows that the more diverse the unlabeled target dataset is, the more performance gain is obtained. In particular, although LFW is similar to MS-Celeb-1M, it can introduce additional diversity in the dataset when augmented with our controllable and guided face synthesis model. Using \emph{unlabeled} IJB-S images as the target data further improves the performance on the IJB-S dataset, which indicates that our model can be applied for boosting face recognition with limited unlabeled samples available.  



\begin{figure}[t]
\vspace{0mm}
\footnotesize
  \resizebox{1\linewidth}{!}{
\begin{tabular}{ c  c  c }

\raisebox{-.5\height}{\includegraphics[scale=0.3]{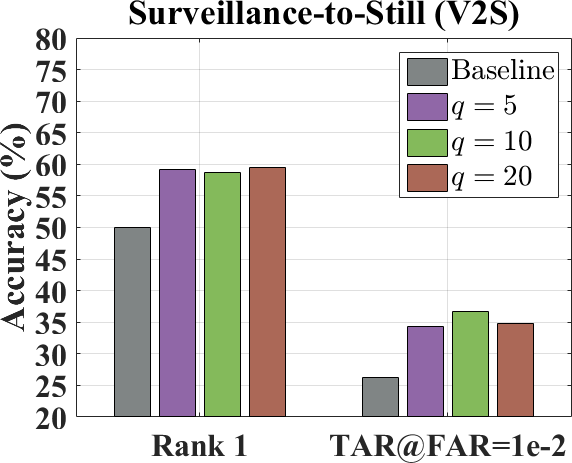}} &
\raisebox{-.5\height}{\includegraphics[scale=0.3]{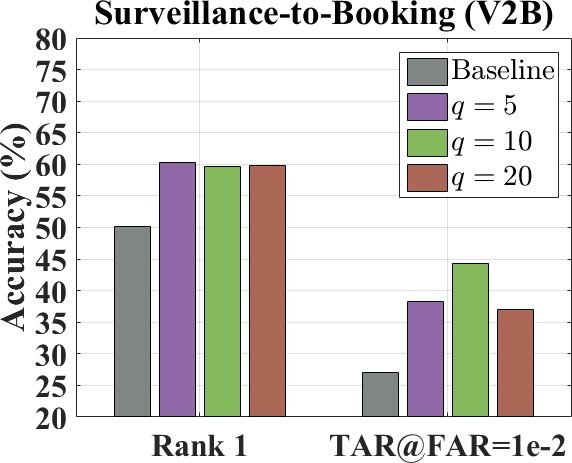}} &
\raisebox{-.5\height}{\includegraphics[scale=0.3]{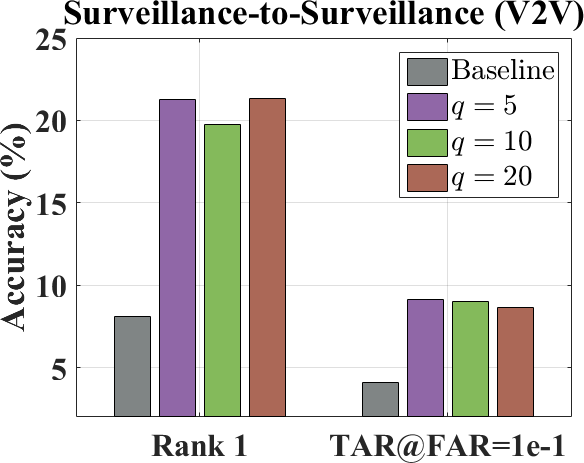}} \\
  \vspace{-4mm}\\ 
  (a)& (b) & (c) \\
\end{tabular}
}
\caption{\small Evaluation results on IJB-S with the different dimensionalities of the style coefficient ($q=5$, $10$, $20$). \textbf{Baseline} refers to the performance of the FR model trained on $0.5$ million labeled samples (a subset of MS-Celeb-1M) without using the proposed face synthesis model. In this experiment, other $3$ models are trained on the $0.5$ million labeled samples with the proposed face synthesis model, which is trained with additional $70$K unlabeled samples from WiderFace.} 
\label{fig:diff_styledim}
\end{figure}

\Paragraph{Effect of the Dimensionality ($q$) of the Style Coefficient.} Fig.~\ref{fig:diff_styledim} shows the recognition performances on IJB-S over the dimensionality of the style coefficient. Fig.~\ref{fig:diff_styledim} shows that the dimensionality of the style coefficient does have significant effects on the recognition performance. The model with $q=10$ performs slightly better in face verification setting, such as V2S and V2B (TAR@FAR=1e-2). The results also indicate that learning manipulation in the low-dimensional subspace is effective and robust for face recognition.   

\begin{figure}[t]
\vspace{0mm}
\footnotesize
  \resizebox{1\linewidth}{!}{
\begin{tabular}{ c  c  c }

\raisebox{-.5\height}{\includegraphics[scale=0.3]{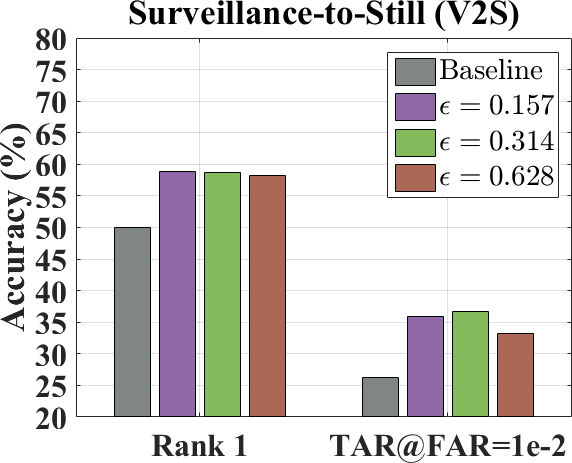}} &
\raisebox{-.5\height}{\includegraphics[scale=0.3]{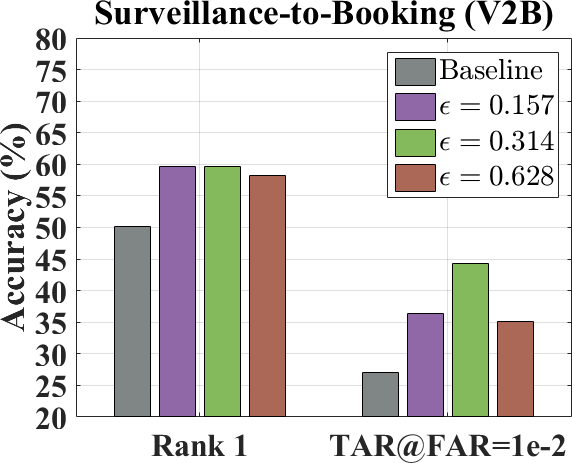}} &
\raisebox{-.5\height}{\includegraphics[scale=0.3]{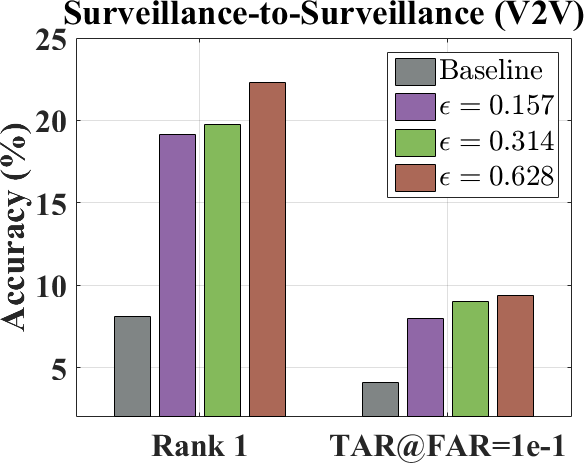}} \\
  \vspace{-4mm}\\ 
  (a)& (b) & (c) \\
\end{tabular}
}
\caption{\small Evaluation results on IJB-S with different perturbation budget values ($\epsilon=0.157$, $0.314$ or $0.628$). \textbf{Baseline} refers to the performance of the FR model trained on $0.5$ million labeled samples (a subset of MS-Celeb-1M) without using the proposed face synthesis model. In this experiment, other $3$ models are trained on the $0.5$ million labeled samples with the proposed face synthesis model, which is trained with additional $70$K unlabeled samples from WiderFace.} 
\label{fig:diff_perturbation}
\end{figure}

\Paragraph{Effect of the Perturbation Budget ($\epsilon$).} We conduct experiments to demonstrate the effect of the perturbation budget $\epsilon$. 
As shown in Fig.~\ref{fig:diff_perturbation}, we can clearly find that a large perturbation budget ($\epsilon=0.628$) leads to a better performance in the protocol of Surveillance-to-Surveillance (V2V) while performs slightly worse in the protocols of Surveillance-to-Still (V2S) and Surveillance-to-Booking (V2B). These observations are not surprising because the large style coefficient perturbation would generate faces with low qualities, which is beneficial for improving generalization to the unconstrained testing scenarios.         



\begin{figure}[t]
\vspace{0mm}
\footnotesize
  \resizebox{1\linewidth}{!}{
\begin{tabular}{ c  c  c }

\raisebox{-.5\height}{\includegraphics[scale=0.3]{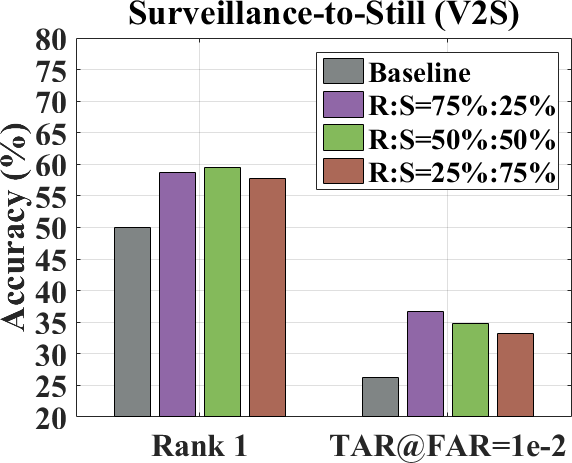}} &
\raisebox{-.5\height}{\includegraphics[scale=0.3]{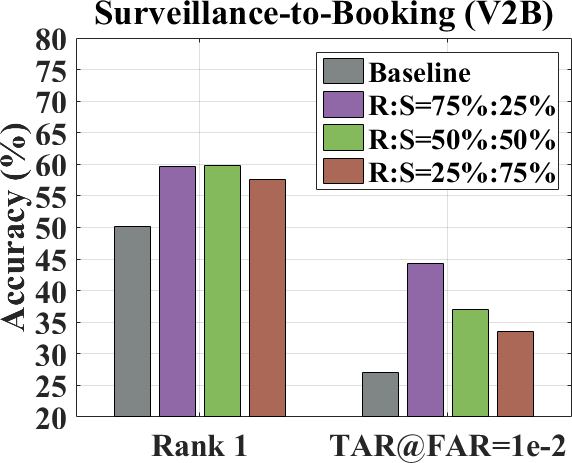}} &
\raisebox{-.5\height}{\includegraphics[scale=0.3]{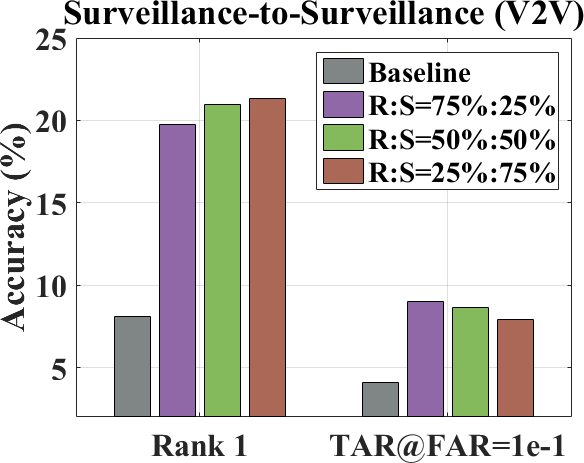}} \\
  \vspace{-4mm}\\ 
  (a)& (b) & (c) \\
\end{tabular}
}
\caption{\small Evaluation results on IJB-S with different ratios of real (R) and synthetic (S) images in each mini-batch (R:S $=75\%:25\%$, $50\%:50\%$ or $25\%:75\%$). \textbf{Baseline} refers to the performance of the FR model trained on $0.5$ million labeled samples (a subset of MS-Celeb-1M) without using the proposed face synthesis model. In this experiment, other $3$ models are trained on the $0.5$ million labeled samples with the proposed face synthesis model, which is trained with additional $70$K unlabeled samples from WiderFace.} 
\label{fig:diff_ratio}
\end{figure}

\Paragraph{Effect of the Ratio of Real and Synthetic Images in Each Mini-batch.} As illustrated in Sec. 3.2 (main paper), we combine the original real images and their corresponding synthesized version as a mini-batch for the FR model training. In this experiment, we further study the ratio of real (R) and synthetic (S) images in each mini-batch. As shown in Fig.~\ref{fig:diff_ratio}, with more synthetic images in each mini-batch (R:S $=25\%:75\%$), the model achieves the best performance in the most challenging Surveillance-to-Surveillance (V2V) protocol (Rank1).

\section{Additional Visualizations}
\begin{figure}[t]
\vspace{0mm}
\footnotesize
  \resizebox{1\linewidth}{!}{
\begin{tabular}{ c  c  c }

\raisebox{-.5\height}{\includegraphics[scale=0.3]{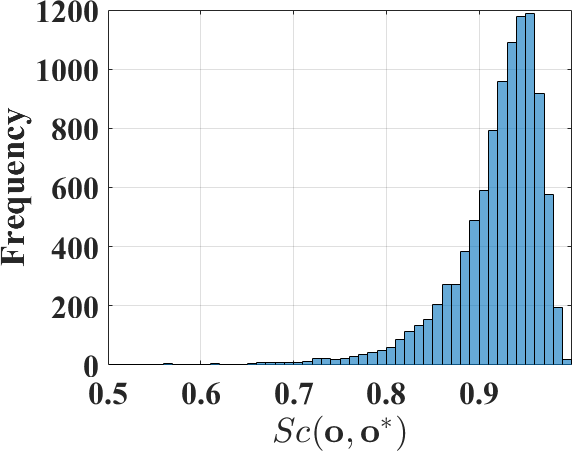}} &
\raisebox{-.5\height}{\includegraphics[scale=0.3]{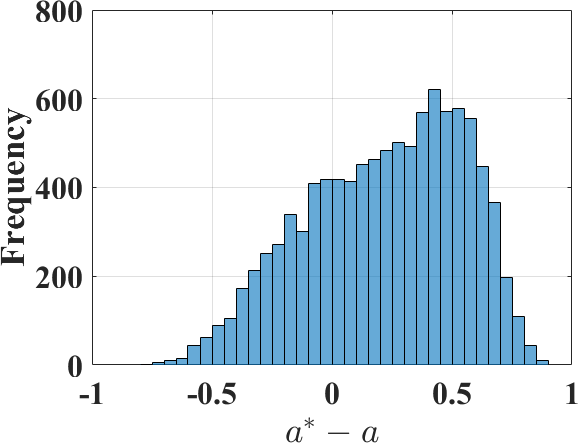}} &
\raisebox{-.5\height}{\includegraphics[scale=0.3]{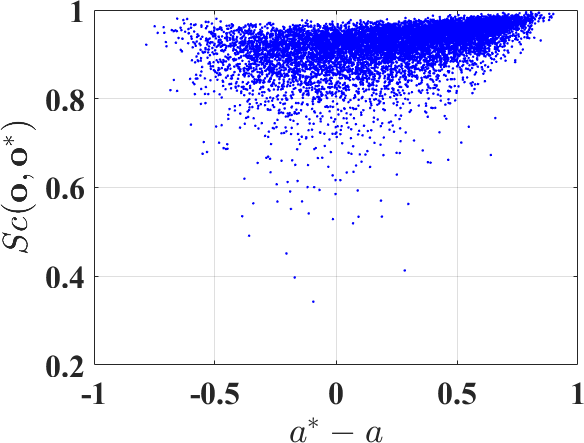}} \\
  \vspace{-3mm}\\ 
  (a)& (b) & (c) \\
\end{tabular}
}
\caption{\small (a) Histogram of the Cosine Similarity between the initial style coefficient $\mathbf{o}$ and its updated one  $\mathbf{o}^*$ with perturbation. (b) Histogram of differences between the magnitude of $\mathbf{o}$ and $\mathbf{o}^{*}$: $a-a^*$, where $a^*=||\mathbf{o}||$, $a=||\mathbf{o}^*||$. (c) Scatter plot showing the correlation between $Sc$ and $a-a^*$.} 
\label{fig:sta_info}
\end{figure}

\Paragraph{The Perturbations in Direction or Magnitude.} In adversarial FR model training, our synthesis model is able to offer two meaningful possibilities to perform style coefficient perturbation: magnitude and direction. To study the perturbation properties (direction or magnitude), we collect the initial style coefficient $\mathbf{o}$ and style perturbation $\boldsymbol\delta^{*}$ of $10$K samples during the FR model training. We first measure the Cosine Similarity $S_C$ (Fig.~\ref{fig:sta_info}~\textcolor{red}{(a)}) between the initial style coefficient $\mathbf{o}$ and the updated one $\mathbf{o}^*=\mathbf{o}+\boldsymbol\delta^*$. Then we present the histogram of the differences  (Fig.~\ref{fig:sta_info}~\textcolor{red}{(b)}) between the magnitude of $\mathbf{o}$ and $\mathbf{o}^{*}$: $a^*-a$, where $a^*=||\mathbf{o}||$, $a=||\mathbf{o}^*||$. Finally, in Fig.~\ref{fig:sta_info}~\textcolor{red}{(c)}, we show the $Sc$ over $(a^*-a)$. As observed in Fig.~\ref{fig:sta_info}, the style coefficient perturbation guided by FR model training indeed leads to the changes of both magnitude and direction of the initial style coefficient, which supports the motivation of our controllable face synthesis model design. More interestingly, the synthesis model attempts to achieve a balance between magnitude and direction in the adversarial-based augmentation process (see Fig.~\ref{fig:sta_info}~\textcolor{red}{(c)}). For example, when the magnitude is decreasing ($(a-a^*)<0$), the model is inclined to generate faces in lower quality but more target styles (lower $Sc$). In contrast, when the magnitude is increasing ($(a-a^*)>0$), the model prefers to generate faces with higher quality but less target style (larger $Sc$).

\begin{figure}[t]
\vspace{0mm}
\newcommand{\tabincell}[2]{\begin{tabular}{@{}#1@{}}#2\end{tabular}}
\footnotesize
\resizebox{1\linewidth}{!}{
\begin{tabular}{ c  c }
\vspace{0.5mm}
{\tiny \tabincell{c}{ Original \\$\mathbf{X}$}}& \raisebox{-.5\height}{\includegraphics[scale=0.28]{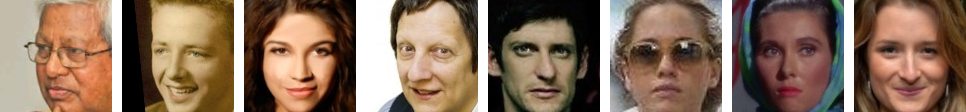}} \\ \vspace{0.5mm}
{\tiny \tabincell{c}{ Synthesized \\$\hat{\mathbf{X}}$}}& \raisebox{-.5\height}{\includegraphics[scale=0.28]{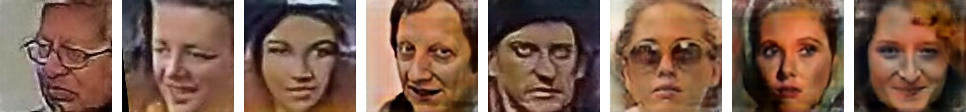}} \\ \vspace{0.5mm}
{\tiny $|\hat{\mathbf{X}}-\mathbf{X}|$} & \raisebox{-.5\height}{\includegraphics[scale=0.28]{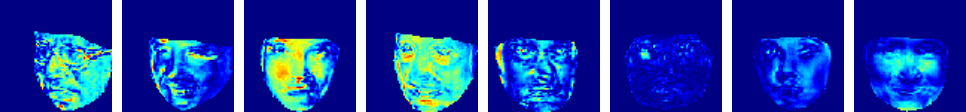}} \\ \vspace{0.5mm}
{\tiny \tabincell{c}{ Guided \\$\mathbf{X}^*$}} & \raisebox{-.5\height}{\includegraphics[scale=0.28]{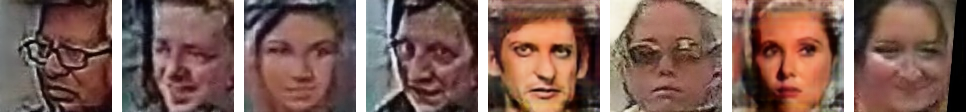}} \\ \vspace{0.5mm}
{\tiny $|\mathbf{X}^*-\mathbf{X}|$}& \raisebox{-.5\height}{\includegraphics[scale=0.28]{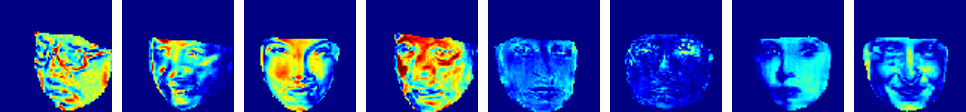}} \\\vspace{2mm}
{\tiny $|\hat{\mathbf{X}}-\mathbf{X}^*|$}& \raisebox{-.5\height}{\includegraphics[scale=0.28]{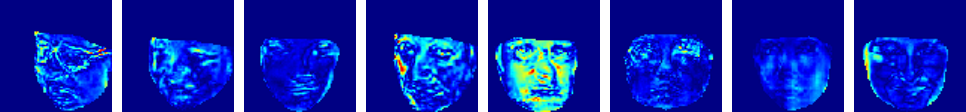}} \\ \vspace{0.5mm}

{\tiny \tabincell{c}{ Original \\$\mathbf{X}$}}& \raisebox{-.5\height}{\includegraphics[scale=0.28]{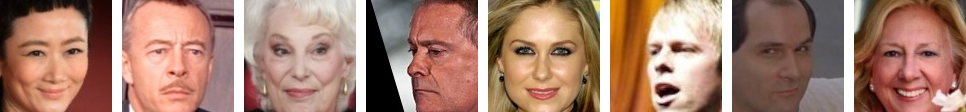}} \\ \vspace{0.5mm}
{\tiny \tabincell{c}{ Synthesized \\$\hat{\mathbf{X}}$}}& \raisebox{-.5\height}{\includegraphics[scale=0.28]{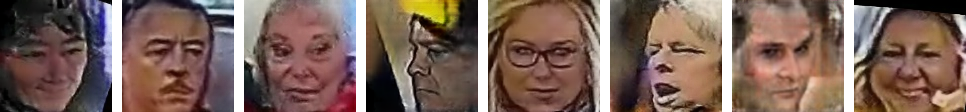}} \\ \vspace{0.5mm}
{\tiny $|\hat{\mathbf{X}}-\mathbf{X}|$} & \raisebox{-.5\height}{\includegraphics[scale=0.28]{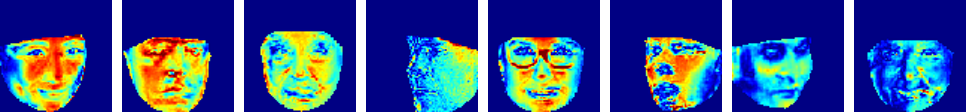}} \\ \vspace{0.5mm}
{\tiny \tabincell{c}{ Guided \\$\mathbf{X}^*$}} & \raisebox{-.5\height}{\includegraphics[scale=0.28]{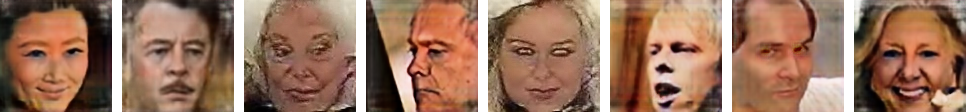}} \\ \vspace{0.5mm}
{\tiny $|\mathbf{X}^*-\mathbf{X}|$}& \raisebox{-.5\height}{\includegraphics[scale=0.28]{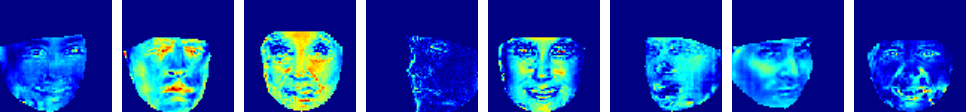}} \\
{\tiny $|\hat{\mathbf{X}}-\mathbf{X}^*|$}& \raisebox{-.5\height}{\includegraphics[scale=0.28]{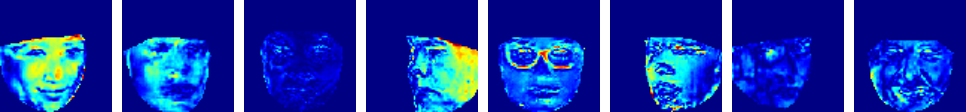}} \\ 
\end{tabular}
}
\caption{ \small Training examples in a mini-batch with our face synthesis model during the FR model training. For each image set, we show the original images $\mathbf{X}$, synthesized results with initial style coefficients, and synthesized results with style perturbations $\mathbf{X}^*$. We additionally show their corresponding error maps: $|\hat{\mathbf{X}}-\mathbf{X}|$, $|\mathbf{X}^*-\mathbf{X}|$ and $|\hat{\mathbf{X}}-\mathbf{X}^*|$.}
\label{fig:adversarial_examples}
  \vspace{0mm}
\end{figure}

\Paragraph{Additional Visualizations of $\mathbf{X}$, $\hat{\mathbf{X}}$ and $\mathbf{X}^{*}$.} In Fig.~\ref{fig:adversarial_examples}, we show the original examples $\mathbf{X}$, synthesized examples with initial style coefficients  $\hat{\mathbf{X}}$ and synthesized examples with style perturbations $\mathbf{X}^*$ in a mini-batch during the FR model training. Additionally, we visualize the pairwise error maps among these $3$ types of data. As shown, the guide from the FR model encourages the face synthesis model to generate images with either increased or decreased target face style.

\clearpage
%
%
\bibliographystyle{splncs04} 
\bibliography{egbib}
\end{document}